\documentclass{article} % For LaTeX2e
\usepackage{iclr2027_conference,times}

\usepackage{geometry}
\usepackage{amsmath,amsfonts,bm}

\def\eqref#1{equation~\ref{#1}}
\def\1{\bm{1}}

\DeclareMathAlphabet{\mathsfit}{\encodingdefault}{\sfdefault}{m}{sl}
\SetMathAlphabet{\mathsfit}{bold}{\encodingdefault}{\sfdefault}{bx}{n}

\usepackage{hyperref}
\usepackage{url}

\usepackage{ulem}
\usepackage{booktabs}
\usepackage{multirow}
\usepackage{graphicx}
\usepackage{adjustbox}
\usepackage{amssymb}
\usepackage[table]{xcolor}
\usepackage{amsmath}
\usepackage{enumitem}
\usepackage{algorithm}
\usepackage{algorithmic}
\usepackage{array}
\usepackage{tabularx}

\title{Blame Errant Words, Not the Canvas: 
\\ Defect-Aware Spatial Credit Assignment for \\ Visual Text Generation}

\iclrfinalcopy

\makeatletter
\def\@maketitle{%
  \vbox{\hsize\textwidth
    \centering
    {\LARGE\sc \@title\par}%
    \vskip 0.28in
    {\normalsize
      \textbf{Sujie Hu}$^{1,6}$,
      \textbf{Huaisong Zhang}$^{2,6}$,
      \textbf{Pujing Yang}$^{3,6}$,
      \textbf{Qian Wang}$^{4,6}$,
      \textbf{Xiaomin Li}$^{5,6}$, \\[0.35em]
      \textbf{Xinhang Leng}$^{6,\dagger}$,
      \textbf{Chen Li}$^{6}$,
      \textbf{Jing Lyu}$^{6}$,
      \textbf{Xiu Li}$^{2}$,
      \textbf{Ling Pan}$^{1,*}$\\[0.85em]
      $^{1}$The Hong Kong University of Science and Technology \quad 
      $^{2}$Tsinghua University \quad
      $^{3}$Zhejiang University \\[0.28em]
      $^{4}$Peking University \quad
      $^{5}$Dalian University of Technology\quad
      $^{6}$Weixin, Tencent \\[0.65em]
      \footnotesize $^{\dagger}$Project lead. \quad $^{*}$Corresponding author.\par}%
    \vskip 0.22in}}
\makeatother

\usepackage{xspace}

\newcommand{\reward}{TextDART\xspace}
\newcommand{\ourrl}{SCANNER\xspace}
\newcommand{\bench}{PolyText-Bench\xspace}
\newcommand{\slogan}{Blame errant words, not the canvas\xspace}

\begin{document}

\maketitle
\vspace{-22pt}
\begin{center}
\normalsize
  \href{https://sujiehu.github.io/textdart-scanner}
       {\textcolor{blue}{\texttt{https://sujiehu.github.io/textdart-scanner}}}
\end{center}

\fancyhf{}
\fancyhead[L]{\small Blame Errant Words, Not the Canvas: Defect-Aware Spatial Credit Assignment for Visual Text Generation}
\fancyfoot[C]{\thepage}
\renewcommand{\headrulewidth}{0.4pt}
% The conference style freezes \headwidth at the old 5.5in measure.
% Recompute it from the current \textwidth whenever a header is built.
\fancyhfoffset{0pt}
\makeatletter
\renewcommand{\headrule}{%
  \hrule\@height\headrulewidth\@width\textwidth
  \vskip-\headrulewidth}
\makeatother

\begin{abstract}
\vspace{-5pt}
Recent reinforcement learning approaches have improved visual text generation, yet generated images still suffer from incorrect wording and malformed glyphs. We identify two coupled limitations: recognition-based rewards often overlook visible glyph defects, while whole-image scalar feedback obscures the uneven quality of individual text regions. These limitations motivate our central principle: \textit{\slogan}, to effectively preserve local quality distinctions. First, we introduce {\reward}, a defect-aware region-targeted reward model that localizes individual text regions and independently evaluates their semantic correctness and glyph quality, providing fine-grained, region-precise feedback. Building on \reward, we propose {\ourrl}, which turns these regional judgments into spatially differentiated policy feedback.
% a reinforcement learning framework with spatial credit assignment for visual text generation. 
It normalizes rewards for each text target across sampled images and casts the resulting regional advantages onto the corresponding latent tokens.
% , blending them with global advantages through a noise-dependent schedule to enable reliable policy updates. 
% We further introduce \textbf{\bench} for evaluating both text-critical generation, where the desired wording is specified, and text-optional generation, where it is not. Extensive experiments demonstrate that our approach substantially outperforms state-of-the-art baselines in text accuracy and glyph fidelity. 
Extensive experiments on existing benchmarks and our newly established \bench demonstrate that our approach achieves state-of-the-art performance in text accuracy and glyph fidelity. 
% Code will be released upon acceptance of the paper.
% \textcolor{red}{simplify bench}
\vspace{-5pt}
\end{abstract}
\section{Introduction}
\label{sec:intro}

% ======================Para. 1======================
% task&tech scope: visual text generation & GRPO
% ======================Para. 1======================
% % What task are we solving
% Diffusion and flow-matching models have achieved remarkable success in image generation, producing high-fidelity and detail-rich visual content. Alongside these advances, visual text generation, which requires synthesizing legible text within generated images, has emerged as a challenging frontier. 
% % Limitation of pretrained models
% Despite their strong general priors, pretrained models still fail frequently at this task, exhibiting missing characters, merged strokes, and hallucinated glyphs. 
% % Recent tech of solutions: GRPO
% Recent works mitigate these failures through reinforcement learning (RL), where rewards computed on sampled images directly optimize text quality during generation. Among RL algorithms, Group Relative Policy Optimization (GRPO) has demonstrated notable stability and scalability in visual alignment.

% t2i visual text generation -> RL post-training -> improvement does not bring reliable visual text rendering -> failure -> why residual visual text failure matters
Text-to-image (T2I) models have made remarkable progress in visual fidelity, compositionality, and prompt adherence~\citep{rombach2022high,podell2024sdxl, flux2024,qwenimage,zhao2026extending}, and are rapidly evolving from generating compelling scenes to producing usable visual content~\citep{wang2026elastic,yu2025omnialpha,He2025ScalingIA,liu2025controllable}. A broad range of visual artifacts (e.g., posters, advertisements) rely on the accurate integration of language and imagery, where a single misspelled word or malformed character can compromise an otherwise compelling image. Reliable visual text generation is therefore a key capability for extending generative models from pictorial synthesis to practical content creation.
% RL and GRPO for t2i
Recent reinforcement learning (RL) post-training methods~\citep{black2024ddpo, xu2023imagereward,fu2026flowawr}, particularly Group Relative Policy Optimization (GRPO)~\citep{flowgrpo,dancegrpo} and its variants~\citep{branchgrpo,mixgrpo,treegrpo,luo2025principled,d2align,e2po,li2026portraitgen,Liu_2026_CVPR,anonymous2026gardo,ma2025stagestablegeneralizablegrpo,ma2026margrpostabilizedgrpoardiffusion,ICLR2026_54c9bfb0,Zhang_2026_CVPR}, have demonstrated 
% promising progress stability and scalability in visual alignment.
promising progress in the stability and scalability of visual alignment.

% visual text generation, but still limitation
Despite this progress, visual text generation remains a challenging frontier~\citep{chen2024textdiffuser2, anytext, yang2023glyphcontrol, ma2024glyphdraw}, where generated images still suffer from incorrect wording and malformed glyphs such as missing characters, merged strokes, and hallucinated shapes.
Consequently, improvements in average image quality can conceal consequential local defects, where a single corrupted word may still render the final artifact unusable. More importantly, in RL post-training, a failure that the reward cannot distinguish is not simply left unmeasured; it may be reinforced as successful behavior~\citep{pan2022rewardmisspecification, skalse2022rewardhacking,fu2026graftood}.

% \begin{figure}[!t]
% \vspace{-20pt}
%   \centering
%     \includegraphics[width=0.99\textwidth]{images/teaser-v4.pdf}
%     % \vspace{-15pt}
%     \caption{\textbf{Image-level text rewards misrank samples and misassign credit.}
% (a) An OCR reward overlooks A's malformed glyph and ranks A over B, so Flow-GRPO reinforces the defect and penalizes B's correct subtitle.
% (b) Rewards alone supervise GRPO through group-computed advantages.
% (c) \reward gives defect-aware regional rewards, and \ourrl compares samples per target, so each region earns its own advantage.
% (d) Ours matches the baseline's final reward with $3.47\times$ GPU-hour efficiency.}
% \label{fig:teaser}
% \label{fig:teaser}
%     \label{fig: tesear}
%     % \vspace{-15pt}
% \end{figure}

% ======================Para. 2======================
% define challenges 
% ======================Para. 2======================
% 2 Key features of visual text generation and current limitation
% However, visual text generation differs from generic alignment in two important ways.
We attribute these persistent failures to two key properties of visual text generation that existing RL methods overlook.
\textbf{(1) Errors are heterogeneous}: a failure may lie in \textit{what} is written (a semantic error) or in \textit{how} it looks (a glyph defect), but most existing methods supervise only the former, leaving deformations invisible to the reward. \textbf{(2) Text is regional}: it occupies localized areas, and a single image may contain multiple text instances of uneven quality, so text quality is not uniformly distributed, yet most methods score the whole image, averaging away local correctness. 
% Principle, ref to title
Accordingly, these observations motivate our central principle: 
% judge each word by its own flaws, or, as we term it, 
\textit{{\slogan}}. 
% Raise two questions
Realizing this principle requires preserving local quality distinctions from evaluation to policy optimization, rather than reducing them to a single image-level advantage as in standard image-level GRPO.
% Off-the-shelf GRPO, however, cannot follow it, 
This raises two questions: \textbf{what should fine-grained rewards measure}, and \textbf{where should reliable credit go}?

\begin{figure}[!h]
% \vspace{-10pt}
  \centering
    \includegraphics[width=0.85\textwidth]{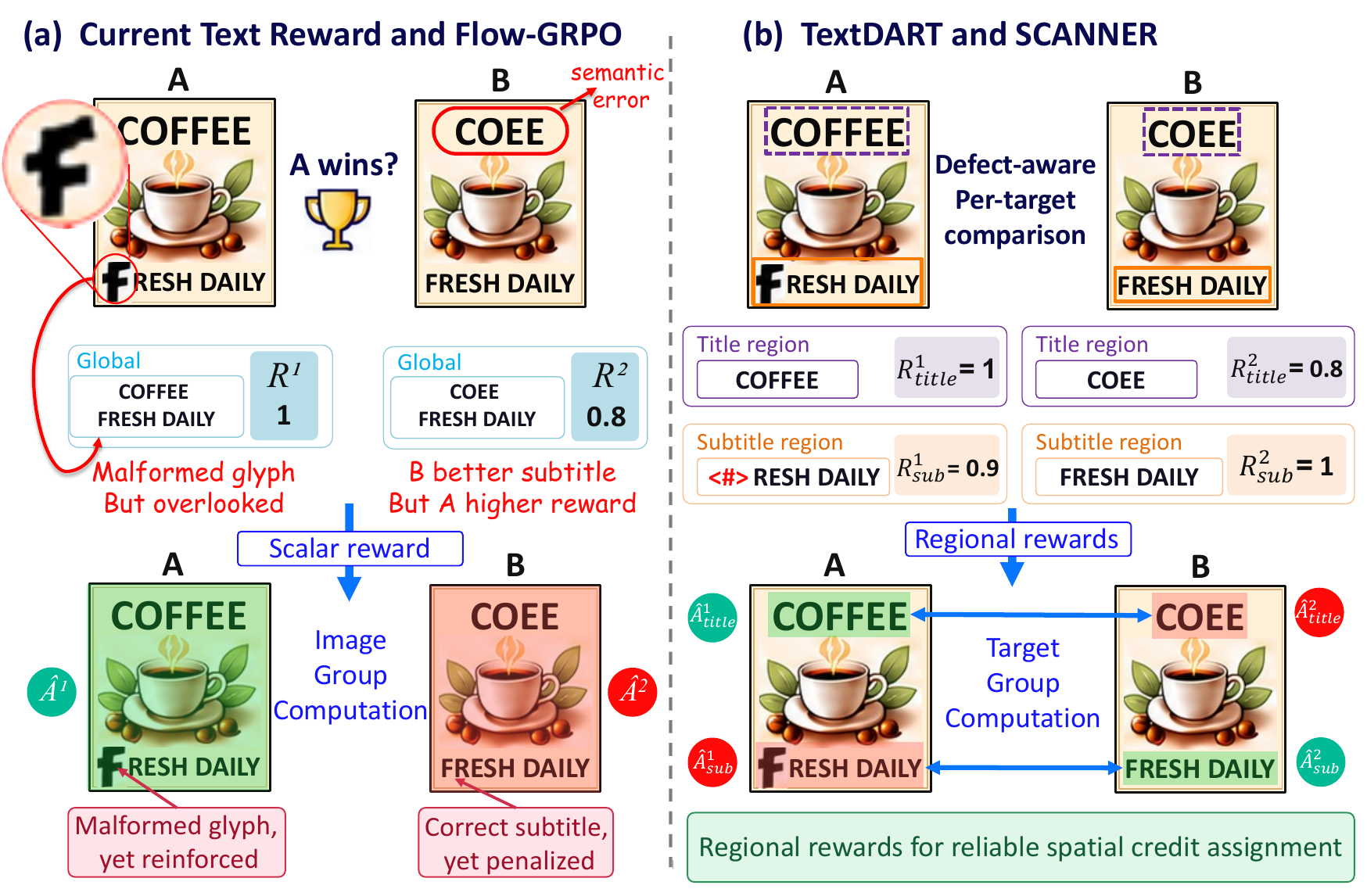}
    \vspace{-10pt}
    \caption{\textbf{Comparison of Our Method Against Current Reward Model and Flow-GRPO.}
(a) A outranks B despite its malformed glyph, and image-group advantages reinforce the defect while penalizing B's correct subtitle.
% (b) \reward scores each region with defect-aware perception (\texttt{<\#>}), and \ourrl computes target-group advantages for finer spatial credit, penalizing only A's malformed subtitle while reinforcing B's correct one (green/red: positive/negative).
(b) \reward provides defect-aware scores for individual regions, and \ourrl computes target-group advantages for spatial credit assignment, penalizing A’s malformed subtitle while reinforcing B’s correct one (green/red: positive/negative).
}
\label{fig:teaser}
    \label{fig: tesear}
    \vspace{-10pt}
\end{figure}

% ======================Para. 3======================
% first challenge->\reward
% ======================Para. 3======================
% Recent method: OCR&VLLM
The first question concerns how rewards are computed, as shown in Fig.\ref{fig: tesear}(a). Existing methods largely rely on OCR~\citep{ppocrv5} or vision-language models to recognize the rendered string and compute an edit-distance-based reward~\citep{li2025textcraftor, cui2025textalign, rahman2025textdiffuserrl}.
% limitations: Hallucination&Lack of defect 
Such recognizers carry strong linguistic priors: malformed glyphs are frequently read as valid words, and obvious stroke merging or omission is hallucinated as normal text. Moreover, the normalized edit distance only reflects semantic correctness and says nothing about glyph quality. 
% One solution: TextPecker
While TextPecker~\citep{textpecker} introduces perceptual signals for defective text, 
% But still limitation: Lack of localization
it operates on the whole image, flattening all text regions into a single score. 
% Our method, \reward, and the pipeline
To obtain fine-grained rewards, we propose a \textbf{D}efect-\textbf{A}ware \textbf{R}egion-\textbf{T}argeted reward model, \textbf{\reward}.
% , trained on train set of \reward-123K.
% Effect: fine-grained rewards
In Fig.\ref{fig: tesear}(b), rather than scoring the whole image, \reward first localizes every text region, then evaluates glyph quality and semantic correctness independently, yielding rewards that are both defect-aware and region-precise.

% ======================Para. 4======================
% second challenge->\ourrl
% ======================Para. 4======================
% Recent method: scalar credit
The second question persists even with \textit{localized, defect-sensitive} feedback in hand. Standard GRPO assigns the same credit to every latent token at every denoising step~\citep{flowgrpo, dancegrpo}.
% limitation: sparse in spatial and temporal
As a result, uniform credit propagation lets flawed regions drag down well-formed ones, and the temporal dynamics of text emergence are ignored. 
% Other solutions: spatial(spatialflow-grpo), temporal(densegrpo,tempflowgrpo)
% But still limitation:
Prior attempts either allocate regional signals~\citep{spatialflow, sdg,ouyang2026region} while ignoring that text layout is undetermined in early steps, or compute rewards at intermediate steps~\citep{densegrpo, truningpoint-grpo,tempflow,zhang2026drm,liu2026temporal}, which is costly and still not region-aware. 
% Our method, \ourrl, and the pipeline
To resolve this, we propose \textbf{\ourrl}, a reinforcement learning
framework for \textbf{S}patial \textbf{C}redit
\textbf{A}ssig\textbf{N}ment with
\textbf{N}ois\textbf{E}-awa\textbf{R}eness.
Utilizing the proposed \reward, a target-group strategy normalizes the rewards per text target and casts them onto the corresponding latent tokens. Along the denoising trajectory, the global advantage is dynamically fused with these regional signals via a noise-dependent schedule, shifting credit from the global view to local focus as text regions gradually resolve, 
% Effect: reliable credit
thereby providing reliable guidance for policy optimization.

% ======================Para. 5======================
% text-optional->\bench
% ======================Para. 5======================
% move to exp
% limitation: lack of measurement in complex scenes
% Existing benchmarks compute metrics against ground-truth strings, leaving text-optional generation---where prompts do not specify the wording---without quantitative measurement in complex scenes. 
% % Our bench, \bench
% Building on the defect-aware reward, we introduce \textbf{\bench}, a novel benchmark for quantitative evaluation of both text-critical and text-optional generation,
% % Effect: comprehensive assess
% thereby offering a standardized means to assess visual text generation in its full diversity.

% ======================Para. 5======================
% contribution
% ======================Para. 5======================
In summary, our main contributions are as follows:
% \begin{itemize}
%     \item We pinpoint the absence of localized, defect-sensitive feedback as a critical barrier to high-fidelity visual text generation. To address this, we introduce \textbf{\reward}, a defect-aware region-targeted reward model trained on our \reward-123K to separately evaluate semantic correctness and glyph quality across individual text regions.

%     \item We propose \textbf{\ourrl}, a spatiotemporal credit assignment framework for flow models. A target-group strategy normalizes the rewards per text target within each group and casts them onto the corresponding latent tokens. Along the denoising trajectory, the global advantage is dynamically fused with these regional signals via a noise-dependent schedule to shift credit from the global view to local focus, thereby providing reliable guidance for policy optimization.

%     \item We establish \textbf{\bench} to benchmark multi-region text-critical and text-optional generation. Extensive experiments demonstrate that our approach achieves superior character accuracy and fidelity, significantly outperforming state-of-the-art baselines.
% \end{itemize}

\begin{itemize}[leftmargin=*, topsep=0pt, parsep=0pt, partopsep=0pt]
\item We pinpoint the absence of localized, defect-sensitive feedback as a critical barrier to visual text generation. To this end, we introduce \textit{\reward}, a defect-aware region-targeted reward model, to disentangle semantic correctness and glyph quality across individual text regions.
\item We propose \textit{\ourrl}, a spatial credit assignment method that normalizes and projects rewards onto target latent tokens, dynamically shifting focus from global advantages to regional signals via a noise-dependent schedule for reliable policy optimization.
% \item We establish \textit{\bench} for multi-region text-optional and text-critical generation. Extensive experiments demonstrate that our method achieves state-of-the-art text accuracy and visual fidelity.
\item Extensive experiments on four existing benchmarks and our newly established \textit{\bench} demonstrate our method achieves state-of-the-art performance in text accuracy and glyph fidelity.
% \textcolor{red}{and our newly established \bench. simplify}
\end{itemize}
\section{Related Work}
\label{sec:related}

\subsection{Visual Text Generation and Evaluation}
\label{sec:related_textgen}
% 生成方法：字形与布局控制 → 文字编码和训练数据增强 → 基于奖励的后训练。
% 评价方法作为奖励：OCR 字符串匹配 → 字形缺陷感知；前者容易忽略视觉缺陷，后者仍缺少整图输入下可定位细粒度的区域奖励。
% 另外评测范围：主要围绕指定文字，未充分覆盖未指定文字内容的 text-optional 场景。
% 本文关注：文字区域级的内容正确性与字形质量，并覆盖 text-critical 和 text-optional 评测。
Visual text generation has been approached by improving rendering through layout or glyph conditioning and glyph-aware training data~\citep{textdiffuser,anytext}, and more recently by post-training pretrained generators with rewards. 
Visual evaluation has increasingly favored fine-grained, spatialized feedback over coarse scalar scores~\citep{liu2025omnidiff,meng2026versusq,Liu2026ArtQuant,han2026vigorbenchfarvisualgenerative}.
The dominant reward is OCR-based string matching~\citep{flowgrpo}, which is blind to visual defects. \citet{textpecker} penalize glyph deformations with perceptual reward but still collapse all text into a single whole-image score, so defects cannot be localized. 
% Meanwhile, existing benchmarks report string accuracy against reference text~\citep{longtextbench,cvtg2k}, leaving text-optional generation in complex scenes unmeasured. 
Building on these observations, we focus on region-level evaluation of both semantic correctness and glyph quality, 
% and extend the evaluation scope to both text-critical and text-optional generation, 
enabling fine-grained feedback.
% \textcolor{red}{bench simplify},

\subsection{RL and Spatial Credit Assignment for Diffusion and Flow Models}
\label{sec:related_spatial}
% 背景：RL 与 GRPO 用于扩散和流模型后训练，但整图标量优势难以区分局部贡献。
% 标量重分配：通过注意力或感知图分配同一优势，能调整局部权重，但不能独立奖惩不同区域。
% 像素级反馈：通过缺陷热图构造局部优势，但相同空间位置未必对应跨样本的同一语义目标。
% 区域级反馈：按语义目标构造优势并更新对应区域，但最终区域在早期高噪声阶段的对应关系可能不可靠。
% 本文关注：按文字目标分配可靠的区域优势，并随去噪过程从全局指导逐步转向局部反馈。
RL post-training aligns diffusion and flow models through preference optimization or GRPO~\citep{diffusion_dpo,flowgrpo,grpo,li2025reneg,shen2026follow}, but a whole-image scalar advantage cannot distinguish local contributions. One remedy reallocates this same advantage according to attention maps~\citep{star} or perceptual allocation maps~\citep{vipo,focusnfix}: spatial weights are adjusted, yet no region can be rewarded or penalized on its own. Pixel-level feedback takes a different route, building local advantages from defect heatmaps~\citep{richhf,imagedoctor,pixelrl,sdg}, but the same spatial position need not correspond to the same semantic target across samples. Region-level feedback instead constructs advantages per semantic target and updates the corresponding regions~\citep{spatialflow,editgrpo}, yet the region-target correspondence is unreliable in early high-noise steps. Building on these observations, we focus on reliable credit assignment per text target that adapts to the denoising process, leading to more effective policy optimization.

\section{Preliminaries}
\label{sec:preliminary}

Diffusion and flow models transport noise to data along Gaussian probability
paths $\mathbf{x}_t=\alpha_t\mathbf{x}_0+\sigma_t\boldsymbol{\epsilon}$,
$\boldsymbol{\epsilon}\sim\mathcal{N}(\mathbf{0},\mathbf{I})$, and admit an ODE
sampler $\mathrm{d}\mathbf{x}_t/\mathrm{d}t=\mathbf{v}_\theta(\mathbf{x}_t,t,c)$
with conditioning prompt $c$~\citep{ddpm,sde,flow1,flow2,lv2025rethinking,chen2026stochastic,he2024diffusion}. Flow-GRPO~\citep{flowgrpo}
enables online RL on such models by replacing the deterministic sampler with a
stochastic SDE counterpart and applying group-relative policy
optimization~\citep{grpo}. For the linear path $\alpha_t=1-t$, $\sigma_t=t$,
its SDE counterpart preserves the ODE marginals under an exact velocity field.
On a grid $0=t_0<\cdots<t_T=1$ with $h_k=t_k-t_{k-1}>0$, the discretized
transition is
\begin{equation}
    \mathbf{x}_{t_{k-1}}\!\sim\!\mathcal{N}(\boldsymbol{\mu}_{\theta,k},n_k^2\mathbf{I}),\quad
    \boldsymbol{\mu}_{\theta,k}=\mathbf{x}_{t_k}-h_k\!\left[
      \mathbf{v}_{\theta,k}
      +\frac{g_{t_k}^{2}}{2t_k}\big(\mathbf{x}_{t_k}+(1-t_k)\mathbf{v}_{\theta,k}\big)\right],
\label{eq:flowgrpo_sampler}
\end{equation}
where $\mathbf{v}_{\theta,k}=\mathbf{v}_\theta(\mathbf{x}_{t_k},t_k,c)$,
$g_{t_k}>0$ is the policy-independent noise strength, and $n_k=g_{t_k}\sqrt{h_k}$
the transition noise.

Given a prompt $c$, the old policy samples $G$ trajectories with terminal
rewards $R_i$, normalized into group-relative advantages
\begin{equation}
    \widehat{A}_i=(R_i-\bar R)/(s_R+\delta),
\label{eq:group_advantage}
\end{equation}
with $\bar R$ and $s_R$ the group mean and standard deviation and $\delta>0$
for numerical stability. Letting
$r_{i,k}(\theta)=p_\theta(\mathbf{x}_{t_{k-1}}^i\mid\mathbf{x}_{t_k}^i,c)/
p_{\theta_{\mathrm{old}}}(\mathbf{x}_{t_{k-1}}^i\mid\mathbf{x}_{t_k}^i,c)$
denote the probability ratio, the policy maximizes
\begin{equation}
    \mathcal{J}(\theta)=\mathbb{E}\!\left[
      \tfrac{1}{GT}\textstyle\sum_{i=1}^{G}\sum_{k=1}^{T}
      \min\!\big(r_{i,k}\widehat{A}_i,\,
      \operatorname{clip}(r_{i,k},1{-}\varepsilon,1{+}\varepsilon)\widehat{A}_i\big)
      -\beta D_{i,k}\right],
\label{eq:flowgrpo_objective}
\end{equation}
where the expectation is over prompts and old-policy trajectories, $\varepsilon$
the clipping threshold, and $D_{i,k}$ the conditional KL to the reference
policy with weight $\beta$, which reduces to a squared difference of transition
means under identical covariances. The same image-level advantage
$\widehat{A}_i$ guides all denoising steps without distinguishing regional
contributions.
\section{Method}
\label{sec:method}

\begin{figure}[!t]
\vspace{-5pt}
  \centering
    \includegraphics[width=1.0\textwidth]{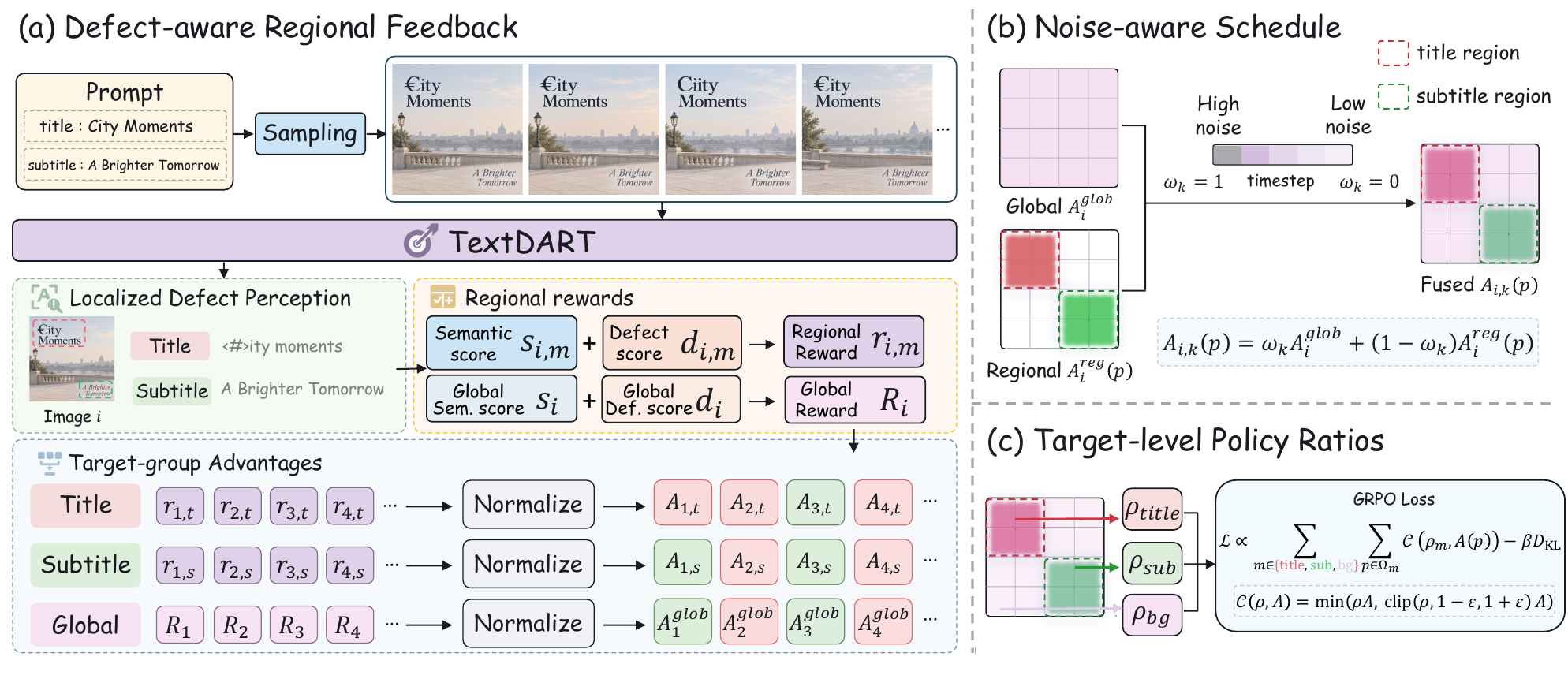}
    \vspace{-20pt}
    \caption{\textbf{Overview of \reward{} and \ourrl{}.}
(a) \reward{} localizes text and evaluates semantic correctness and defects. Rewards for the same target are normalized across samples to construct spatial advantages.
(b) A noise-aware schedule shifts credit from global to regional advantages as denoising proceeds.
(c) Policy ratios are pooled within each text target and the background, enabling spatially aligned GRPO updates.}
    \label{fig: pipeline}
    \vspace{-5pt}
\end{figure}

% \begin{figure*}[!t]
% % \vspace{-10pt}
%   \centering
%     \includegraphics[width=\textwidth]{images/vis-reward.pdf}
%     \vspace{-20pt}
%     \caption{\textbf{Localization and defect-aware recognition under challenging visual conditions.} \reward accurately localizes and reads text while marking \texttt{<\#>} as malformed glyphs, whereas PP-OCRv5 hallucinates incorrect content and TextPecker misses or mismarks defects.}
%     \label{fig: vis-reward}
%   % \vspace{-10pt}
% \end{figure*}

Given a prompt that specifies the target text, we aim to generate images with accurate content and defect-free text. However, recognition-based rewards may overlook text defects due to hallucination, while image-level rewards erase regional quality differences. Reliable optimization therefore requires attributing credit to individual errant words rather than the whole image. To this end, as shown in Fig.~\ref{fig: pipeline}, we introduce \reward for defect-aware text recognition and localization, and \ourrl for spatial credit assignment from its regional feedback.

\subsection{\reward: \underline{D}efect-\underline{A}ware \underline{R}egion-\underline{T}argeted Rewards}
\label{sec:method-reward}
\paragraph{Localized Defect Perception.}
To perceive text defects together with their locations, we fine-tune Qwen3-VL-8B on the training set of \reward{}-123K (detailed in \ref{supp:textdart-123K}) to jointly predict text quadrilaterals and transcriptions with structural-anomaly markers:
\begin{equation}
\{(q_{i,o},t_{i,o})\}_{o=1}^{O_i}=\mathcal{F}_{\mathrm{text}}(I_i),
\label{eq:textdart_prediction}
\end{equation}
where $\mathcal{F}_{\mathrm{text}}$ denotes \reward, $I_i$ the $i$-th sampled image, $O_i$ the number of regions originally detected by $\mathcal{F}_{\mathrm{text}}$, and $q_{i,o}$ and $t_{i,o}$ the quadrilateral and defect-aware transcription of region $o$. The model takes only the image as input, keeping visual assessment independent of the target text, which is then used for matching and scoring. 
% As shown in Fig.~\ref{fig: vis-reward}, \reward provides more accurate defect-aware localization and recognition than TextPecker and PP-OCRv5.

\paragraph{Regional Rewards.}
A single image-level score fails to distinguish text regions of varying quality. As shown in Fig.~\ref{fig: pipeline}(a), we therefore match regional transcriptions to target text through many-to-many matching and compute rewards separately (both detailed in \ref{supp:algorithm}). Regions matched to the same target are merged, while targets covered by one region are also merged consistently throughout the rollout. This yields a shared set of $M$ merged target texts $\bar y_m$, each with region $\bar q_{i,m}$ and transcription $\bar t_{i,m}$ in image $i$ when present. Motivated by \citet{textpecker}, the regional reward $r_{i,m}$ combines the semantic correctness score $s_{i,m}$ and the defect-free score $d_{i,m}$, with higher scores indicating better quality.
\begingroup
\small
\begin{equation}
 s_{i,m}=1-\operatorname{GNED}(\bar t_{i,m},\bar y_m),\quad
 d_{i,m}=\left[1-\kappa\frac{a(\bar t_{i,m})}{L(\bar t_{i,m})}\right]_0^1,\quad
 r_{i,m}=(1-w_d)s_{i,m}+w_d d_{i,m},
\label{eq:regional_reward}
\end{equation}
\endgroup
where $\operatorname{GNED}$ denotes the generalized normalized edit distance, $a$ the weighted defect count, $L$ the transcription length, $\kappa$ the defect-penalty scale, $w_d$ the mixing weight, and $[\cdot]_0^1$ the clipping operator on $[0,1]$. A target absent from image $i$ receives $r_{i,m}=0$, and regions matched to no target are merged into a single extra entry $r_{i,\mathrm{ext}}=d_{i,\mathrm{ext}}$ computed on their concatenated transcription, which is counted as a reward entry only when such extra text is present.

\subsection{\ourrl: \underline{S}patial \underline{C}redit \underline{A}ssig\underline{n}ment with \underline{n}ois\underline{e}-awa\underline{r}eness}
\label{sec:method-rl}

\ourrl converts the regional rewards of \reward into token-level credit through three components: target-group advantages that compare the same text target across samples, a noise-aware schedule that shifts credit from global to regional signals along denoising, and target-level policy ratios that keep updates consistent within each target.

\paragraph{Target-group Advantages.}

% Why we use target-group
Standard GRPO gives all image tokens one advantage, so flawed and well-formed regions receive identical credit. Regional rewards require a comparison unit: averaging within an image reduces to image-level comparison, while fixed latent positions are sensitive to layout variation. We therefore group regions aligned with the same merged target across images generated from one prompt and normalize rewards within each target group. Since these regions render identical content, the advantage isolates generation quality(Fig.~\ref{fig: pipeline}(a)):
% how to do
\begin{equation}
 A_{i,m}=\frac{r_{i,m}-\bar r_m}{\max(\sigma_m^{r},\delta)},\quad
 \bar r_m=\operatorname{mean}_{i'}r_{i',m},\quad
 \sigma_m^{r}=\operatorname{std}_{i'}r_{i',m},
\label{eq:target_advantage}
\end{equation}
where $A_{i,m}$ denotes the advantage for target $m$ in image $i$, $\bar r_m$ and $\sigma_m^{r}$
the reward mean and sample standard deviation over the $G$ images $i'$ sampled for the same prompt,
and $\delta>0$ the standard-deviation floor. The extra entry $r_{i,\mathrm{ext}}$ is normalized in the same way with a clean reference of $1$ for its absence, and its advantage is assigned only to existing extra-text regions.

% To cast the per-region target advantage into the token-level objective, we project the merged regions $\bar q_{i,m}$ onto the latent grid and normalize each contribution by its support size, since a uniform broadcast would give larger text more weight simply for occupying more tokens:
% \begin{equation}
%  A_i^{\mathrm{reg}}(p)=\frac{\lambda N}{|\mathcal M_i|}
%  \sum_{m\in\mathcal M_i}\frac{[p\in\Omega_{i,m}]}{|\Omega_{i,m}|}A_{i,m},
% \label{eq:advantage_projection}
% \end{equation}
% where $A_i^{\mathrm{reg}}(p)$ denotes the regional advantage at latent position $p$, $N$ the number of latent tokens, $\Omega_{i,m}$ the token support of $\bar q_{i,m}$, $\mathcal M_i\subseteq\{1,\ldots,M\}\cup\{\mathrm{ext}\}$ the indices of targets present in image $i$, and $\lambda$ the regional scale, with $|\cdot|$ denoting cardinality and $[\cdot]$ an indicator. Overlapping tokens belong to the smaller region, and background tokens receive zero advantage, as do all tokens when $\mathcal M_i$ is empty. Each target thus contributes equally regardless of its size.

% To cast the per-region target advantage into the token-level objective, we
% project each merged region $\bar q_{i,m}$ onto the latent grid and normalize its
% contribution by support size, so that every target receives equal total credit
% regardless of how many tokens it occupies:
To incorporate per-target regional advantages into the token-level objective,
we project each merged region $\bar q_{i,m}$ onto the latent grid and
normalize its contribution by its token-support size, removing
area-dependent weighting at the advantage-map normalization stage:
\begin{equation}
 A_i^{\mathrm{reg}}(p)=\frac{\lambda N}{|\mathcal M_i|}
 \sum_{m\in\mathcal M_i}\frac{[p\in\Omega_{i,m}]}{|\Omega_{i,m}|}A_{i,m},
\label{eq:advantage_projection}
\end{equation}
where $A_i^{\mathrm{reg}}(p)$ is the regional advantage at latent position $p$,
$N$ is the number of latent tokens, $\Omega_{i,m}$ is the token support of
$\bar q_{i,m}$, $\mathcal M_i\subseteq\{1,\ldots,M\}\cup\{\mathrm{ext}\}$
indexes target entries with nonempty spatial support in image $i$,
either observed directly or inferred from peer images, and $\lambda$ is the regional scale,
with $|\cdot|$ and $[\cdot]$ denoting cardinality and the indicator function.
Tokens in overlapping regions are assigned to the smaller region, while
background tokens---and all tokens when $\mathcal M_i$ is empty---receive zero
advantage.

\paragraph{Noise-aware Schedule.}
Regional advantages rely on text locations, which are undetermined in early denoising steps, so credit should shift from global feedback toward regional signals as denoising proceeds. 
% For the global branch, rather than averaging regional rewards, which weights each target equally regardless of its length, we score the concatenated transcription and targets jointly, rather than averaging independently computed regional rewards:
For the global branch, we score the concatenated transcription and targets
jointly instead of averaging independently computed regional rewards:
\begin{equation}
 R_i=(1-w_d)s(\operatorname{concat}_o t_{i,o},\operatorname{concat}_j y_j)
       +w_d d(\operatorname{concat}_o t_{i,o}),
\label{eq:global_text_reward}
\end{equation}
where $R_i$ denotes the global reward, $y_j$ the $j$-th original target, and $s$ and $d$ the scoring functions in Eq.~\ref{eq:regional_reward}, with $\operatorname{concat}$ denoting space-separated concatenation in detected-region order for transcriptions and in target order for targets.
As shown in Fig.~\ref{fig: pipeline}(b), we then mix the two advantages at each step, with the global weight decaying as the transition noise decreases:
\begin{equation}
\begin{aligned}
 \omega_k&=\frac{\widetilde n_k-\min_{l\in\mathcal S}\widetilde n_l}
 {\max_{l\in\mathcal S}\widetilde n_l-\min_{l\in\mathcal S}\widetilde n_l},\\
 A_{i,k}(p)&=\omega_k A_i^{\mathrm{glob}}+(1-\omega_k)A_i^{\mathrm{reg}}(p),
\end{aligned}
\label{eq:noise_aware_fusion}
\end{equation}
where $\omega_k$ denotes the global mixing weight, $\widetilde n_k$ the running minimum in generation order of the transition noise standard deviation $n_k=g_{t_k}\sqrt{h_k}$ from Eq.~\ref{eq:flowgrpo_sampler}, $A_i^{\mathrm{glob}}$ the global advantage from Eq.~\ref{eq:group_advantage}, $A_{i,k}(p)$ the fused advantage, and $\mathcal S$ the optimized SDE steps indexed by $k$ and $l$. If the noise range is numerically zero, $\omega_k$ is set to 0.5.

\paragraph{Target-level Policy Ratios.}
With token-level advantages in place, the ratio granularity still matters: token-level ratios allow separate clipping decisions within the same text region, whereas an image-level ratio couples unrelated regions. As shown in Fig.~\ref{fig: pipeline}(c), to keep updates consistent within each merged target while retaining regional differences, we pool the Guard-adjusted ratio exponent~\citep{grpo-guard} over its token support:
\begin{equation}
\begin{aligned}
 z_{i,k,p}&=n_k(\ell_{i,k,p}^{\theta}-\ell_{i,k,p}^{\mathrm{old}})
                  +\frac{\Delta_{i,k,p}}{2n_k},\\
 \rho_{i,k,m}&=\exp\!\left(\frac{1}{|\Omega_{i,m}|}
                      \sum_{p\in\Omega_{i,m}}z_{i,k,p}\right),
\end{aligned}
\label{eq:target_policy_ratio}
\end{equation}
where $z_{i,k,p}$ denotes the adjusted token exponent, $\rho_{i,k,m}$ the ratio for merged target $m$, including the extra-text entry $\mathrm{ext}$, $\ell_{i,k,p}^{\theta}$ and $\ell_{i,k,p}^{\mathrm{old}}$ the channel-averaged Gaussian log densities of the same sampled transition under the current and old policies, and $\Delta_{i,k,p}$ the channel-averaged squared difference between their transition means. Background tokens form a separate group, and gradients propagate through the pooled exponent.

Text rewards do not directly constrain the surrounding scene. We therefore regularize toward a frozen reference with greater background weight while maximizing the clipped policy objective:
\begin{equation}
 \mathcal J_{\mathrm{SCANNER}}(\theta)=\mathbb E\!\left[
 \frac{1}{G|\mathcal S|}\sum_{i=1}^{G}\sum_{k\in\mathcal S}
 \left\{\frac{1}{Nh_k}\sum_{p=1}^{N}
 \mathcal C\big(\rho_{i,k}(p),A_{i,k}(p)\big)-\beta D_{i,k}^{\mathrm{KL}}\right\}\right],
\label{eq:castflow_objective}
\end{equation}
where $\mathcal J_{\mathrm{SCANNER}}$ denotes the training objective, and $\rho_{i,k}(p)$ the pooled ratio assigned to token $p$. The clipped surrogate is $\mathcal C(\rho,A)=\min\{\rho A,\operatorname{clip}(\rho,1-\varepsilon,1+\varepsilon)A\}$ for ratio $\rho$ and advantage $A$, with clipping threshold $\varepsilon$ and regularization weight $\beta$ from Eq.~\ref{eq:flowgrpo_objective}. The penalty $D_{i,k}^{\mathrm{KL}}$ denotes the text/background-weighted KL divergence between the current and reference transition distributions, which reduces to the weighted squared velocity difference under equal transition noise. The expectation is over prompts and old-policy trajectories, and $1/h_k$ denotes the Guard step-size factor.
\section{Experiments}
\label{sec:exp}

\subsection{Experimental Setup.}
\paragraph{Implementation Details.}

\begin{figure*}[!t]
\vspace{-5pt}
  \centering
    \includegraphics[width=0.9\textwidth]{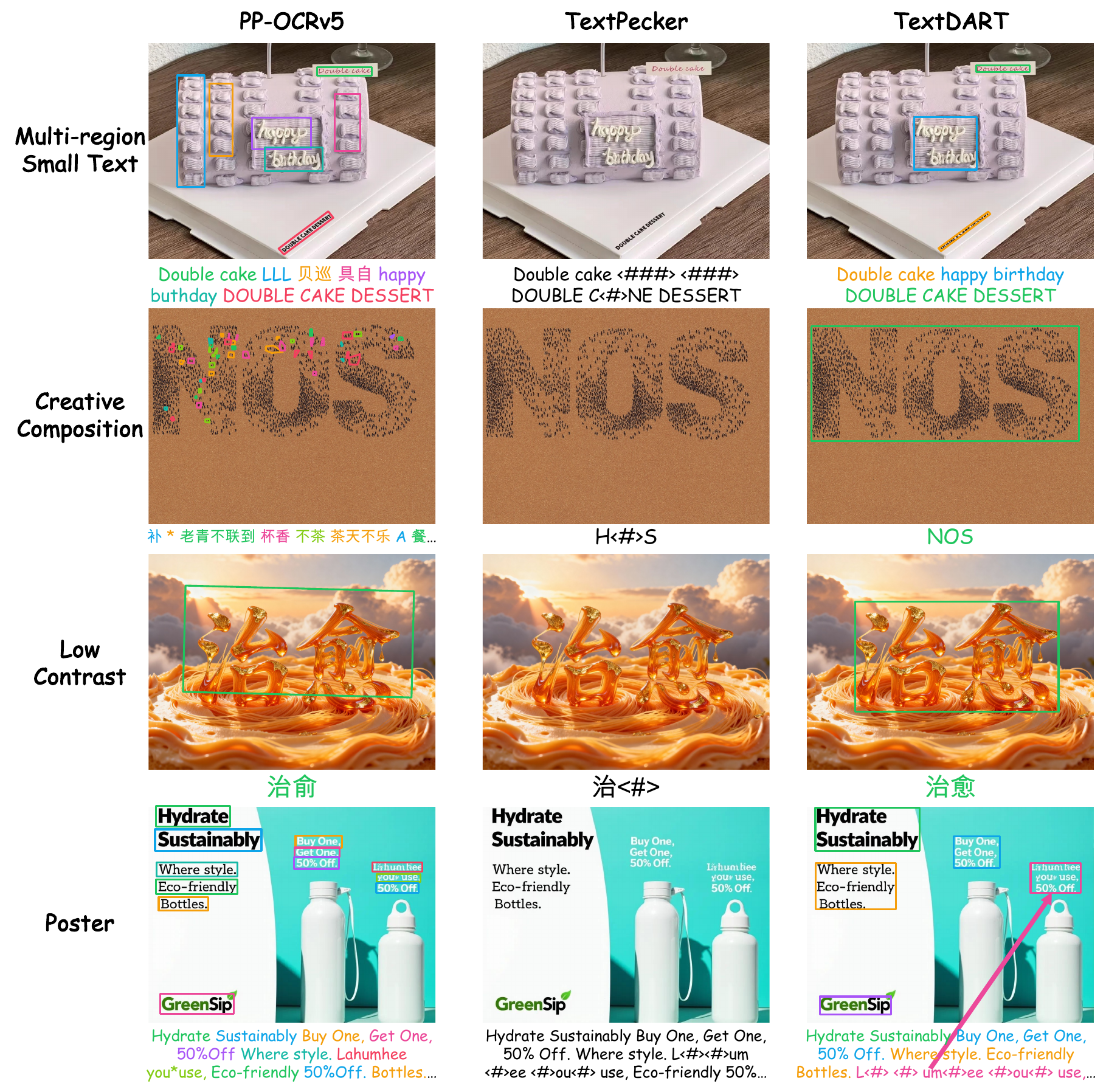}
    \vspace{-10pt}
    \caption{\textbf{Localization and Defect-aware Recognition under Challenging Visual Conditions.} \reward accurately localizes and reads text while marking \texttt{<\#>} as malformed glyphs, whereas PP-OCRv5 hallucinates incorrect content and TextPecker misses or mismarks defects.}
    \label{fig: vis-reward-extrem}
  \vspace{-5pt}
\end{figure*}

\begin{figure*}[!t]
% \vspace{-10pt}
  \centering
    \includegraphics[width=1.0\textwidth]{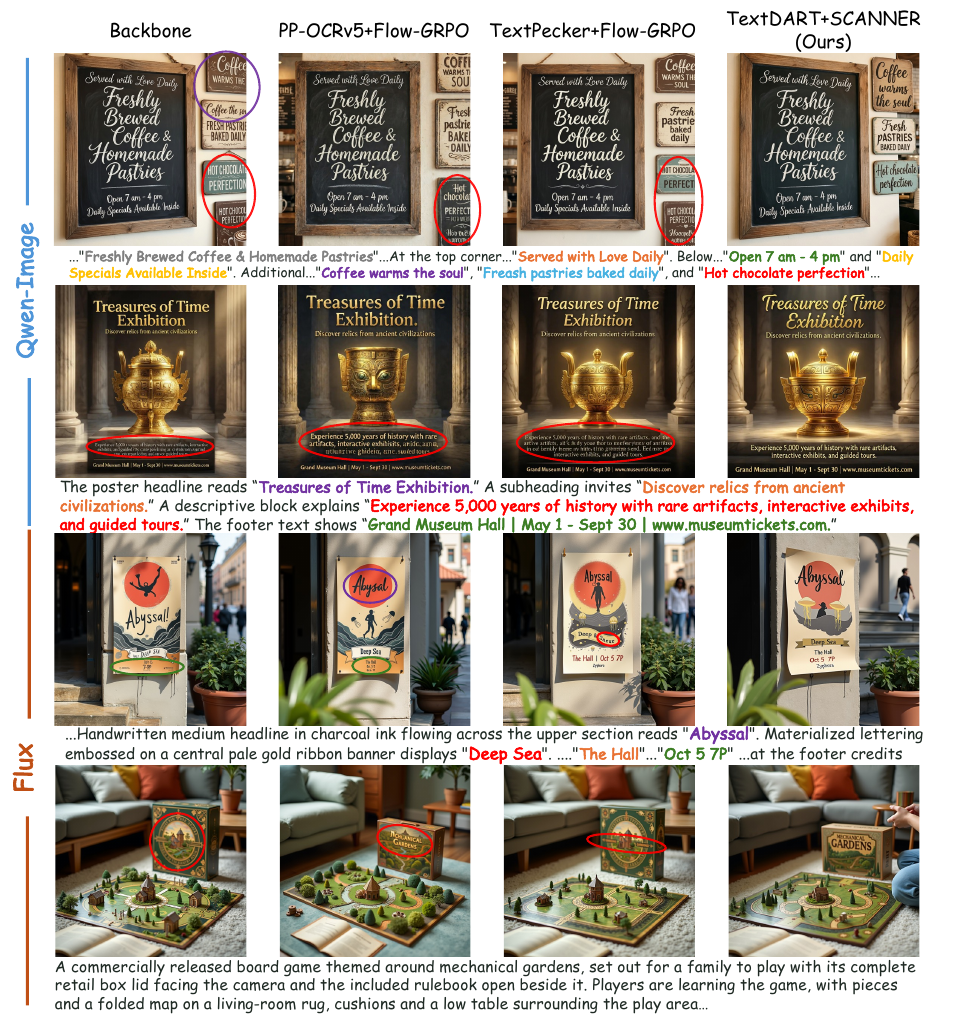}
    \vspace{-10pt}
    \caption{\textbf{Qualitative Comparison of \reward with \ourrl against Baselines.} Target strings are color-coded in the prompts, and circles highlight text errors of the baselines. Our method renders every target correctly while preserving the background and aesthetics of the backbone, and produces legible, well-styled spontaneous text even when no text is prescribed.}
    \label{fig: vis-main}
    \vspace{-15pt}
  % \end{subfigure}
\end{figure*}

% We first evaluate the recognition and localization of \reward, with qualitative comparison
% in Fig.~\ref{fig: vis-reward-extrem} 
% and detailed quantitative results presented in \ref{supp:exp-reward}. 
We perform RL post-training with LoRA on FLUX.1-dev~\citep{flux2024} for English and on Qwen-Image-2512~\citep{qwenimage} for both English and Chinese, using multi-region text prompts with specified targets. We compare against the pretrained backbones and Flow-GRPO~\citep{flowgrpo} equipped with PP-OCRv5~\citep{ppocrv5} or TextPecker~\citep{textpecker} rewards, on LongText-Bench~\citep{longtextbench}, OneIG-Bench~\citep{oneig}, CVTG-2k~\citep{cvtg2k} (English only), GenTextEval-Bench~\citep{textpecker}, and our \bench. Further experimental details are provided in Appendix~\ref{supp:exp-detail}.

\paragraph{\bench.}
Existing benchmarks prescribe target strings and thus cover only text-critical generation. Yet models often render text spontaneously, and defective text, once present, degrades the entire image, making text quality an important aspect of image quality in text-optional scenes. \bench therefore covers both text-critical and text-optional settings with 600 prompts, evenly split between English and Chinese and between the two settings. All metrics are computed by \reward from the full image: critical semantic (CS) and glyph (CG) scores average over requested targets, with missing targets scored zero, while optional glyph quality (OG) averages regional glyph scores over images with detected text, excluding text-free outputs. More details are provided in \ref{supp:bench}.

\subsection{Evaluation of \reward.}
Fig.~\ref{fig: vis-reward-extrem} compares \reward with PP-OCRv5 and TextPecker under challenging visual conditions. \reward preserves valid content in small and low-contrast text that TextPecker mismarks as defective, identifies the malformed glyph in flawed WordArt that both baselines decode as a normal character, and reads the creative ``NOS'' composition as one coherent region that PP-OCRv5 fragments. Detailed quantitative comparison in \ref{supp:exp-reward} further shows that \reward achieves the best results in recognition, defect perception, and localization.

\begin{table*}[t]
\centering
\vspace{-10pt}
\caption{\textbf{Quantitative Comparison of Visual Text Generation Methods.}
($\uparrow$/$\downarrow$: higher/lower is better.)
Best in \textbf{bold}; ``$-$'' denotes no reward model or unsupported language.}
\label{tab:main_results}
\small
\setlength{\tabcolsep}{3pt}
\renewcommand{\arraystretch}{1.15}
\begin{adjustbox}{max width=\textwidth}
\begin{tabular}{@{}ll*{12}{c}@{}}
\toprule
\multirow{2}{*}{Reward} & \multirow{2}{*}{RL Algorithm}
& \shortstack{LongText-Bench}
& \multicolumn{4}{c}{OneIG-Bench}
& \multicolumn{2}{c}{CVTG-2k}
& \multicolumn{2}{c}{GenTextEval-Bench}
& \multicolumn{3}{c}{\bench} \\
\cmidrule(lr){3-3}
\cmidrule(lr){4-7}
\cmidrule(lr){8-9}
\cmidrule(lr){10-11}
\cmidrule(lr){12-14}
& & TextScore $\uparrow$ & TextScore $\uparrow$ & ED $\downarrow$ & CR $\uparrow$ & WAC $\uparrow$ & WordAcc $\uparrow$ & NED $\uparrow$ & Qua $\uparrow$ & Sem $\uparrow$ & CS $\uparrow$ & CG $\uparrow$ & OG $\uparrow$ \\
\midrule
\multicolumn{14}{l}{\textbf{English Text Generation}} \\
\addlinespace[2pt]
\multicolumn{14}{c}{\textit{FLUX.1-dev}} \\
\addlinespace[2pt]
$-$ & $-$ & 0.620 & 0.543 & 85.31 & 0.138 & 0.379 & 0.503 & 0.694 & 0.661 & 0.352 & 0.426 & 0.464 & 0.281 \\
OCRv5 & Flow-GRPO & 0.769 & 0.751 & 64.55 & 0.220 & 0.506 & 0.686 & 0.818 & 0.838 & 0.589 & 0.703 & 0.726 & 0.365 \\
TextPecker & Flow-GRPO & 0.801 & 0.770 & 63.19 & 0.221 & 0.533 & 0.772 & 0.879 & 0.836 & 0.582 & 0.789 & 0.809 & 0.405 \\
\textbf{\reward} & \textbf{\ourrl} & \textbf{0.827} & \textbf{0.858} & \textbf{51.06} & \textbf{0.309} & \textbf{0.598} & \textbf{0.834} & \textbf{0.914} & \textbf{0.839} & \textbf{0.613} & \textbf{0.882} & \textbf{0.905} & \textbf{0.653} \\
\addlinespace[5pt]
\multicolumn{14}{c}{\textit{Qwen-Image-2512}} \\
\addlinespace[2pt]
$-$ & $-$ & 0.952 & 0.986 & 24.54 & 0.461 & 0.895 & 0.844 & 0.922 & 0.944 & 0.787 & 0.887 & 0.899 & 0.252 \\
OCRv5 & Flow-GRPO & 0.947 & 0.989 & 19.58 & 0.509 & 0.890 & 0.832 & 0.914 & 0.946 & 0.795 & 0.876 & 0.876 & 0.273 \\
TextPecker & Flow-GRPO & 0.960 & 0.988 & 22.14 & 0.484 & 0.891 & 0.870 & 0.936 & 0.956 & 0.801 & 0.908 & 0.918 & 0.269 \\
\textbf{\reward} & \textbf{\ourrl} & \textbf{0.972} & \textbf{0.995} & \textbf{13.44} & \textbf{0.585} & \textbf{0.919} & \textbf{0.878} & \textbf{0.939} & \textbf{0.970} & \textbf{0.844} & \textbf{0.953} & \textbf{0.970} & \textbf{0.496} \\
\midrule
\multicolumn{14}{l}{\textbf{Chinese Text Generation}} \\
\addlinespace[2pt]
\multicolumn{14}{c}{\textit{Qwen-Image-2512}} \\
\addlinespace[2pt]
$-$ & $-$ & 0.966 & 0.983 & 13.06 & 0.554 & 0.851 & $-$ & $-$ & 0.955 & 0.853 & 0.931 & 0.945 & 0.198 \\
OCRv5 & Flow-GRPO & 0.965 & 0.986 & 10.88 & 0.584 & 0.850 & $-$ & $-$ & 0.959 & 0.855 & 0.931 & 0.935 & 0.224 \\
TextPecker & Flow-GRPO & 0.974 & 0.985 & 11.83 & 0.574 & 0.853 & $-$ & $-$ & 0.962 & 0.866 & 0.937 & 0.945 & 0.190 \\
\textbf{\reward} & \textbf{\ourrl} & \textbf{0.980} & \textbf{0.992} & \textbf{8.56} & \textbf{0.691} & \textbf{0.856} & $-$ & $-$ & \textbf{0.975} & \textbf{0.892} & \textbf{0.956} & \textbf{0.970} & \textbf{0.403} \\
\bottomrule
\end{tabular}
\end{adjustbox}
\vspace{-10pt}
\end{table*}

\subsection{Qualitative Evaluation.}
Fig.~\ref{fig: vis-main} compares \reward with \ourrl against the pretrained backbone and Flow-GRPO with OCRv5 or TextPecker rewards. On text-critical prompts, baseline errors concentrate on secondary regions (the side panels in the first row and the ribbon banner in the third). The gap is largest on long text: only ours renders the full descriptive block in the second row legibly, while preserving the layout and lighting of the poster. In the text-optional case (last row), ours spontaneously renders a clean, well-styled title on the box lid, while the baselines garble or deform it.

\subsection{Quantitative Evaluation.}
As shown in Tab.~\ref{tab:main_results}, our method achieves the best results on all metrics across both backbones and languages. The largest gains appear on FLUX.1-dev (CR from $0.138$ to $0.309$, WordAcc from $0.503$ to $0.834$), and clear margins hold on the stronger Qwen-Image-2512 (CR from $0.461$ to $0.585$ in English, ED from $13.06$ to $8.56$ in Chinese), where the OCRv5 reward instead degrades several metrics below the base model and TextPecker's image-level reward brings only modest gains. A user study further shows that human evaluators prefer our method over the baselines in \ref{app:user_study}.

\vspace{-.1in}
\subsection{Ablation Study}
\paragraph{Contributions of Reward Components.}
% Reward计算：消融reward每个分数的有效性
% 语义标量 → 加入缺陷评分 → region-mean 标量 → CastFlow
% Tab.~\ref{tab:reward_credit_ablation} augments the reward and the optimizer step by step. Semantic rewards alone bring large gains over the base model on FLUX, and defect scoring improves further. Collapsing regional rewards into an image-level scalar instead degrades Flow-GRPO (ED from $56.88$ to $68.69$), as flawed regions drag down well-formed ones under a broadcast value. \ourrl achieves the best results with the same rewards (ED $51.06$, WordAcc $0.834$), confirming that fine-grained rewards require spatially aligned credit assignment.

Tab.~\ref{tab:reward_credit_ablation} augments the reward and the optimizer step by step. Semantic rewards alone bring large gains over the base model, and defect scoring improves further. Collapsing regional rewards into an image-level scalar instead degrades most Chinese metrics of Flow-GRPO, as flawed regions drag down well-formed ones under a broadcast value. \ourrl achieves the best results with the same rewards (ED $13.44$, WordAcc $0.878$), confirming that fine-grained rewards require spatially aligned credit assignment.

\begin{table*}[t]
\vspace{-.1in}
\centering
\caption{\textbf{Ablation of Reward Components on Qwen-Image.} Sem and Def denote semantic and defect-aware scoring, respectively, and Reg denotes equal-weight aggregation of region-wise rewards into an image-level scalar. The best results in each language are marked in \textbf{bold}, and ``$-$'' marks unsupported languages.}
\label{tab:reward_credit_ablation}
\small
\setlength{\tabcolsep}{3pt}
\renewcommand{\arraystretch}{1.12}
\begin{adjustbox}{max width=\textwidth}
\begin{tabular}{@{}cccl*{12}{c}@{}}
\toprule
\multirow{2}{*}{Sem}
& \multirow{2}{*}{Def}
& \multirow{2}{*}{Reg}
& \multirow{2}{*}{RL Algorithm}
& \shortstack{LongText-Bench}
& \multicolumn{4}{c}{OneIG-Bench}
& \multicolumn{2}{c}{CVTG-2k}
& \multicolumn{2}{c}{GenTextEval-Bench}
& \multicolumn{3}{c}{\bench} \\
\cmidrule(lr){5-5}
\cmidrule(lr){6-9}
\cmidrule(lr){10-11}
\cmidrule(lr){12-13}
\cmidrule(lr){14-16}
& & &
& TextScore $\uparrow$
& TextScore $\uparrow$
& ED $\downarrow$
& CR $\uparrow$
& WAC $\uparrow$
& WordAcc $\uparrow$
& NED $\uparrow$
& Qua $\uparrow$
& Sem $\uparrow$
& CS $\uparrow$ & CG $\uparrow$ & OG $\uparrow$ \\
\midrule
\multicolumn{16}{l}{\textit{English Text Generation}} \\
\addlinespace[2pt]
-- & $-$ & $-$ & Base
& 0.952
& 0.986 & 24.54 & 0.461 & 0.895
& 0.844 & 0.922
& 0.944 & 0.787 & 0.887 & 0.899 & 0.252 \\

$\checkmark$ & $-$ & $-$ & Flow-GRPO
& 0.958
& 0.988 & 22.50 & 0.500 & 0.892
& 0.848 & 0.923
& 0.950 & 0.798 & 0.903 & 0.916 & 0.264 \\

$\checkmark$ & $\checkmark$ & $-$ & Flow-GRPO
& 0.960
& 0.991 & 18.82 & 0.510 & 0.901
& 0.846 & 0.922
& 0.945 & 0.803 & 0.904 & 0.913 & 0.287 \\

$\checkmark$ & $\checkmark$ & $\checkmark$ & Flow-GRPO
& 0.961 & 0.993 & 19.41 & 0.541 & 0.917
& 0.867 & 0.934
& 0.952 & 0.801 & 0.914 & 0.924 & 0.289 \\

\textbf{$\checkmark$} & \textbf{$\checkmark$} & \textbf{$\checkmark$} & \textbf{\ourrl}
& \textbf{0.972}
& \textbf{0.995} & \textbf{13.44} & \textbf{0.585} & \textbf{0.919}
& \textbf{0.878} & \textbf{0.939}
& \textbf{0.970} & \textbf{0.844} & \textbf{0.953} & \textbf{0.970} & \textbf{0.496} \\

\midrule
\multicolumn{16}{l}{\textit{Chinese Text Generation}} \\
\addlinespace[2pt]
-- & $-$ & $-$ & Base
& 0.966
& 0.983 & 13.06 & 0.554 & 0.851
& $-$ & $-$
& 0.955 & 0.853 & 0.931 & 0.945 & 0.198 \\

$\checkmark$ & $-$ & $-$ & Flow-GRPO
& 0.971
& 0.985 & 11.84 & 0.569 & 0.849
& $-$ & $-$
& 0.962 & 0.866 & 0.935 & 0.946 & 0.213 \\

$\checkmark$ & $\checkmark$ & $-$ & Flow-GRPO
& 0.975
& 0.988 & 10.48 & 0.621 & 0.854
& $-$ & $-$
& 0.963 & 0.867 & 0.944 & 0.951 & 0.222 \\

$\checkmark$ & $\checkmark$ & $\checkmark$ & Flow-GRPO
& 0.972 & 0.987 & 11.43 & 0.610 & 0.852
& $-$ & $-$
& 0.962 & 0.867 & 0.941 & 0.949 & 0.218 \\

\textbf{$\checkmark$} & \textbf{$\checkmark$} & \textbf{$\checkmark$} & \textbf{\ourrl}
& \textbf{0.980}
& \textbf{0.992} & \textbf{8.56} & \textbf{0.691} & \textbf{0.856}
& $-$ & $-$
& \textbf{0.975} & \textbf{0.892} & \textbf{0.956} & \textbf{0.970} & \textbf{0.403} \\

\bottomrule
\end{tabular}
\end{adjustbox}
\end{table*}

\begin{table*}[t]
\centering
\vspace{-5pt}
\caption{\textbf{Ablation of Credit Assignment on Qwen-Image.} Global-only and regional-only advantages against their noise-aware combination. The best results in each language are marked in \textbf{bold}, and ``$-$'' marks unsupported languages.}
\label{tab:advantage_mixing_comparison}
\small
\setlength{\tabcolsep}{3pt}
\renewcommand{\arraystretch}{1.12}
\begin{adjustbox}{max width=\textwidth}
\begin{tabular}{@{}l*{12}{c}@{}}
\toprule
\multirow{2}{*}{Advantage Assignment}
& \shortstack{LongText-Bench}
& \multicolumn{4}{c}{OneIG-Bench}
& \multicolumn{2}{c}{CVTG-2k}
& \multicolumn{2}{c}{GenTextEval-Bench}
& \multicolumn{3}{c}{\bench} \\
\cmidrule(lr){2-2}
\cmidrule(lr){3-6}
\cmidrule(lr){7-8}
\cmidrule(lr){9-10}
\cmidrule(lr){11-13}
& TextScore $\uparrow$ & TextScore $\uparrow$ & ED $\downarrow$ & CR $\uparrow$ & WAC $\uparrow$ & WordAcc $\uparrow$ & NED $\uparrow$ & Qua $\uparrow$ & Sem $\uparrow$ & CS $\uparrow$ & CG $\uparrow$ & OG $\uparrow$ \\
\midrule
\multicolumn{13}{l}{\textit{English Text Generation}} \\
\addlinespace[2pt]
Global-only & 0.960 & 0.991 & 18.82 & 0.510 & 0.901 & 0.846 & 0.922 & 0.945 & 0.803 & 0.904 & 0.913 & 0.287 \\
Regional-only & 0.966 & 0.992 & 17.26 & 0.551 & 0.902 & 0.857 & 0.928 & \textbf{0.970} & 0.840 & 0.951 & 0.961 & 0.368 \\
\textbf{Noise-aware schedule} & \textbf{0.972} & \textbf{0.995} & \textbf{13.44} & \textbf{0.585} & \textbf{0.919} & \textbf{0.878} & \textbf{0.939} & \textbf{0.970} & \textbf{0.844} & \textbf{0.953} & \textbf{0.970} & \textbf{0.496} \\
\midrule
\multicolumn{13}{l}{\textit{Chinese Text Generation}} \\
\addlinespace[2pt]
Global-only & 0.975 & 0.988 & 10.48 & 0.621 & 0.854 & $-$ & $-$ & 0.963 & 0.867 & 0.944 & 0.951 & 0.222 \\
Regional-only & 0.978 & 0.991 & 9.43 & 0.652 & 0.855 & $-$ & $-$ & 0.973 & 0.891 & \textbf{0.956} & 0.967 & 0.293 \\
\textbf{Noise-aware schedule} & \textbf{0.980} & \textbf{0.992} & \textbf{8.56} & \textbf{0.691} & \textbf{0.856} & $-$ & $-$ & \textbf{0.975} & \textbf{0.892} & \textbf{0.956} & \textbf{0.970} & \textbf{0.403} \\
\bottomrule
\end{tabular}
\end{adjustbox}
\end{table*}

\begin{figure*}[!t]
\vspace{-2pt}
  \centering
    \includegraphics[width=1.0\textwidth]{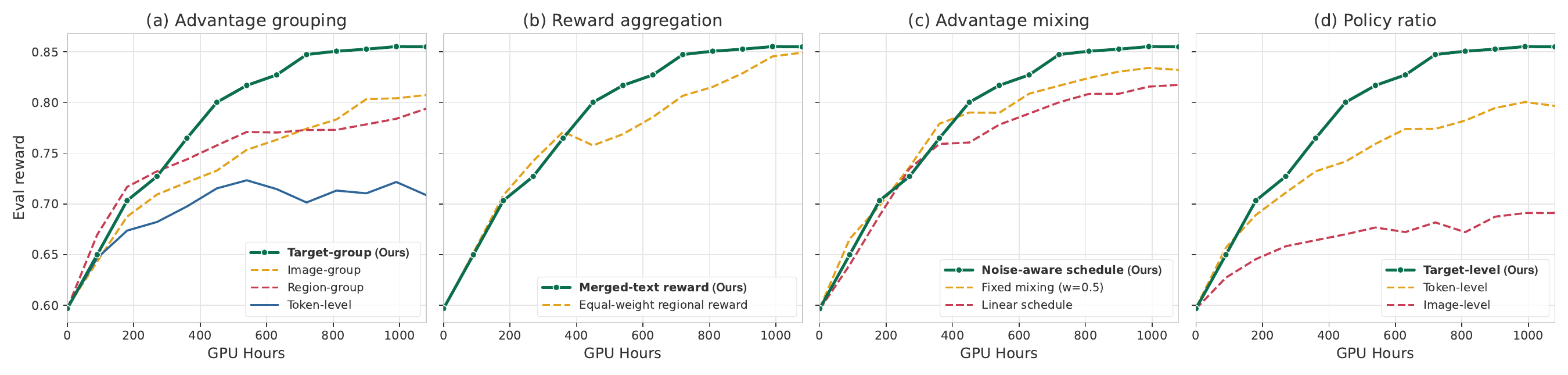}
    \vspace{-10pt}
    \caption{\textbf{Component Ablations of \ourrl.} Evaluation reward throughout training when varying (a) advantage grouping, (b) global reward aggregation, (c) global--regional advantage mixing, and (d) policy ratio granularity. Each curve varies a single component with all other settings fixed, and our design (solid green) attains the highest final reward in every comparison.}
    \label{fig: curve-main}
    \vspace{-10pt}
\end{figure*}

\vspace{-.1in}
\paragraph{Ablation of Credit Assignment.}
% We ablate the components of \ourrl following the credit-assignment pipeline. For advantage grouping in Fig.~\ref{fig: curve-main}(a), token-level normalization stalls early with oscillating rewards, and the region-group variant, which broadcasts the mean regional reward to merged text regions only, plateaus below image-group since all targets share identical credit, while target-group advantages converge the highest by isolating generation quality per target. For the global reward in Fig.~\ref{fig: curve-main}(b), the equal-weight regional mean lags behind the merged-text reward throughout training, as per-region normalization amplifies errors in short text while diluting those in long text. For advantage mixing, Tab.~\ref{tab:advantage_mixing_comparison} shows that regional-only assignment consistently beats global-only, while their noise-aware combination surpasses both, indicating that localized credit is the primary driver and global feedback still helps before text locations become reliable. Fig.~\ref{fig: curve-main}(c) further shows that fixed and linear schedules plateau or lag throughout, while the noise-aware schedule aligns the shift with the transition noise and converges the highest. Finally, for the policy ratio in Fig.~\ref{fig: curve-main}(d), the image-level ratio stalls around $0.69$, as one clipping decision lets defective regions block updates for the whole image, and token-level ratios recover only partially, while target-level ratios converge the highest. 
We ablate \ourrl along its credit-assignment pipeline. For advantage grouping in Fig.~\ref{fig: curve-main}(a), token-level normalization oscillates, region grouping plateaus due to identical regional credit, and target grouping converges best by isolating per-target quality. For the global reward in Fig.~\ref{fig: curve-main}(b), equal-weight regional averaging lags behind merged-text reward because per-region normalization overweights short-text errors while diluting long-text ones. Tab.~\ref{tab:advantage_mixing_comparison} shows that regional credit dominates global credit, while their noise-aware fusion performs best by leveraging global feedback before locations stabilize. Thus, the noise-aware schedule in Fig.~\ref{fig: curve-main}(c) outperforms fixed and linear alternatives. Finally, target-level policy ratios yield the highest final reward, whereas image-level clipping allows defective regions to suppress global updates and token-level ratios recover only partially (Fig.~\ref{fig: curve-main}(d)).
Additional ablation results are provided in \ref{supp:ablation}.
\section{Conclusion}
\label{sec:conclusion}
In this work, we address RL post-training for visual text generation, where recognition-based rewards overlook glyph defects and image-level credit obscures uneven text-region quality. Following the principle of blaming errant words rather than the whole canvas, we introduce \reward, a defect-aware region-targeted reward model, and \ourrl, which assigns spatially aligned credit to each text target. Experiments show our method achieves better text accuracy and glyph fidelity than state-of-the-art baselines.

\bibliography{iclr2027_conference}
\bibliographystyle{iclr2027_conference}

\newpage
\appendix
\section{Appendix}
\subsection{Algorithmic Details}
\label{supp:algorithm}
\paragraph{Many-to-many Matching.}
Detected regions and prompt targets rarely stand in one-to-one correspondence: a target may span several regions, one region may contain several targets, and a region may render text that no target requests. The natural one-to-one baseline assigns each region to its single most similar target,
\begin{equation}
 j^*_{i,o}=\arg\max_{1\leq j\leq K}\left[1-\operatorname{GNED}(t_{i,o},y_j)\right],
\label{eq:greedy_region_match}
\end{equation}
where $K$ denotes the number of targets in the prompt and $j^*_{i,o}$ the target selected for region $o$ of image $i$. This single-target-per-region rule fails in the last two cases. First, an extra-text region is forced onto its closest target, corrupting that target's transcription and reward with content the prompt never requested. Second, a region covering several targets can be credited to only one of them. If a region reads \textit{SUNRISE CAFE} for the targets \textit{SUNRISE} and \textit{CAFE}, greedy matching scores the region against one target with semantic score $0.5$ (reward $0.75$ at $w_d=0.5$), while the other target receives zero reward despite being correctly rendered. We therefore match regions and targets in a many-to-many manner, allowing one region to correspond to multiple targets, multiple regions to contribute to one target, and regions without any counterpart to remain unmatched as extra text.

We match readable text as sequences of Chinese characters and lowercased alphanumeric runs after removing anomaly markers, and region $o$ is initially associated with
\begin{equation}
\begin{aligned}
 \mathcal J^{(0)}_{i,o}
 &=\mathcal C_{i,o}\cup\mathcal F_{i,o}
 \cup\{j:\mathcal U(t_{i,o})\cap\mathcal U(y_j)\neq\varnothing,\ \operatorname{GNED}(t_{i,o},y_j)\leq0.5\}\\
 &\quad\cup\{j:\operatorname{GNED}(t_{i,o},y_j)\leq0.4\},
\end{aligned}
\label{eq:initial_region_correspondence}
\end{equation}
where $\mathcal C_{i,o}$ collects the targets occurring contiguously within the region (\emph{coverage}), $\mathcal F_{i,o}$ the shortest target containing the region as a contiguous fragment, and $\mathcal U(\cdot)$ the readable units. The last two terms tolerate spelling and recognition errors that defeat exact containment. Regions that remain unmatched are attached to their nearest matched neighbor when their transcription occurs in, or flanks, that neighbor's target, recovering line-broken fragments, while regions without readable units stay unmatched.

Merging targets independently per image would make the reward of the same target index refer to different content across samples. We therefore pool coverage relations over all $G$ images of the prompt and merge through connected components,
\begin{equation}
 \mathcal H=\bigcup_{i=1}^{G}\bigcup_{o=1}^{O_i}\bigl\{\{j,j'\}:j,j'\in\mathcal C_{i,o},\ j<j'\bigr\},\qquad
 \{\mathcal T_m\}_{m=1}^{M}=\operatorname{CC}\bigl(\{1,\ldots,K\},\mathcal H\bigr),
\label{eq:shared_target_partition}
\end{equation}
where $\mathcal T_m$ contains the original-target indices of merged target $m$ and $\operatorname{CC}$ denotes connected components over the target-pair edges $\mathcal H$. Only coverage creates merging edges, while fragment and similarity matches associate regions without forcing a merge. If one image renders two targets in a single region and another renders them separately, both are thus scored against the same concatenated target, which the target-group advantages of \ourrl require. The ablation in Fig.~\ref{fig: curve-supp}(a) confirms that this matching improves training over the greedy variant.

\paragraph{Detailed Reward Computation}
Given the shared partition, each merged target collects its detections in image $i$,
\begin{equation}
\begin{aligned}
 \mathcal O_{i,m}&=\{o:\mathcal J_{i,o}\cap\mathcal T_m\neq\varnothing\},
 &\bar y_m&=\operatorname{join}_{j\in\mathcal T_m}y_j,\\
 \mathcal Q_{i,m}&=\bigcup_{o\in\mathcal O_{i,m}}q_{i,o},
 &\bar t_{i,m}&=\operatorname{join}_{o\in\mathcal O_{i,m}}t_{i,o},
\end{aligned}
\label{eq:region_target_merge}
\end{equation}
where $\mathcal J_{i,o}$ denotes the final matched-target indices of region $o$, $\mathcal O_{i,m}$ the detections contributing to merged target $m$, and $\mathcal Q_{i,m}$ their spatial union. The operator $\operatorname{join}$ concatenates in target order for $\bar y_m$ and in spatial reading order for $\bar t_{i,m}$, skipping transcriptions that duplicate a nested detection.

Following ~\citet{textpecker}, the semantic score compares the collected transcription with the merged target through GNED, which tokenizes text into Chinese characters and lowercased non-Chinese runs and solves a unit-level assignment,
\begin{equation}
 \operatorname{GNED}(x,y)=
 \frac{\displaystyle\min_{\pi\in\Pi_{n_x,n_y}}
 \sum_{(v,w)\in\pi}
 \frac{\operatorname{ED}(x_v,y_w)}{\max(|x_v|,|y_w|)}
 +|n_x-n_y|}{\max(n_x,n_y)},
\label{eq:regional_gned_detail}
\end{equation}
where $x$ and $y$ denote the tokenized transcription and target with $n_x$ and $n_y$ units, $\Pi_{n_x,n_y}$ the one-to-one unit matchings of cardinality $\min(n_x,n_y)$, and $\operatorname{ED}$ the character-level edit distance, with unmatched units incurring unit cost. The minimum is computed with the Hungarian algorithm. Since GNED is insensitive to the order of whole units, it measures textual agreement rather than reading order.

Recognizable text can still contain malformed glyphs, so semantic agreement alone is insufficient. The defect-free score counts the structural-anomaly markers with severity-dependent weights,
\begin{equation}
 a(x)=n_{\mathrm{char}}(x)+3\,n_{\mathrm{word}}(x)+5\,n_{\mathrm{block}}(x)+n_{\mathrm{bare}}(x),\qquad
 L(x)=L_{\mathrm{read}}(x)+a(x),
\label{eq:regional_defect_detail}
\end{equation}
where $n_{\mathrm{char}}$, $n_{\mathrm{word}}$, and $n_{\mathrm{block}}$ count the character-, word-, and block-level markers \texttt{<\#>}, \texttt{<\#\#\#>}, and \texttt{<\#\#\#\#\#>}, $n_{\mathrm{bare}}$ the remaining bare \texttt{\#} symbols, and $L_{\mathrm{read}}$ the non-whitespace characters after marker removal. Substituting into Eq.~\ref{eq:regional_reward} gives $d_{i,m}=d(\bar t_{i,m})$, with $d(x)=0$ when $L(x)=0$, so a transcription consisting entirely of defect markers receives zero defect-free score at the training setting $\kappa=1$.

Extra text has no prescribed content against which to measure semantic correctness, while a missing target has a reference but no detected location. We retain both cases distinctly. Unmatched detections form the extra entry,
\begin{equation}
 \mathcal O_{i,\mathrm{ext}}=\{1,\ldots,O_i\}\setminus\bigcup_{m=1}^{M}\mathcal O_{i,m},\qquad
 (\bar q_{i,\mathrm{ext}},\bar t_{i,\mathrm{ext}})=\Bigl(\bigcup_{o\in\mathcal O_{i,\mathrm{ext}}}q_{i,o},\ \operatorname{join}_{o\in\mathcal O_{i,\mathrm{ext}}}t_{i,o}\Bigr),
\label{eq:extra_text_region}
\end{equation}
and an unobserved target borrows its location from peer images of the same prompt,
\begin{equation}
 \bar q_{i,m}=\begin{cases}
 \mathcal Q_{i,m},&\mathcal O_{i,m}\neq\varnothing,\\
 \operatorname{bbox}\!\left(\bigcup_{i'\neq i}\mathcal Q_{i',m}\right),&\mathcal O_{i,m}=\varnothing,
 \end{cases}
\label{eq:missing_text_region}
\end{equation}
where $\operatorname{bbox}$ denotes the smallest enclosing rectangle, and the inferred region excludes tokens already occupied by observed regions when projected onto the latent grid. The reward entries are then
\begin{equation}
 r_{i,m}=\begin{cases}
 (1-w_d)s_{i,m}+w_d d_{i,m},&\mathcal O_{i,m}\neq\varnothing,\\
 0,&\mathcal O_{i,m}=\varnothing,
 \end{cases}
 \qquad
 r_{i,\mathrm{ext}}=d(\bar t_{i,\mathrm{ext}})\ \ \text{only if}\ \mathcal O_{i,\mathrm{ext}}\neq\varnothing,
\label{eq:regional_reward_cases}
\end{equation}
with the spatially available entries
\begin{equation}
 \mathcal M_i=\{m:\bar q_{i,m}\neq\varnothing\}\cup\{\mathrm{ext}:\mathcal O_{i,\mathrm{ext}}\neq\varnothing\}.
\label{eq:active_text_regions}
\end{equation}
A target absent from every sampled image has no inferred region and is excluded from spatial assignment. These distinctions let omitted targets and malformed extra text receive localized feedback without penalizing the entire canvas.

\paragraph{\ourrl Algorithm}
We summarize the complete training procedure in Algorithm~\ref{alg:castflow}. For each prompt with specified target texts, we first sample a group of $G$ images with the SDE sampler and record the transition noise $n_k$ of each optimized step. \reward then localizes text regions with defect-aware transcriptions, which are matched to the target texts and scored into regional rewards $r_{i,m}$ and a global reward $R_i$. Subsequently, the regional rewards are normalized within each target group and projected onto the latent grid, while the global reward is normalized within the prompt, and the two advantages are fused at each step with the noise-aware weight $\omega_k$. Finally, we pool the Guard-adjusted ratio exponent over the token support of each target and update the policy with the clipped objective and the background-weighted KL regularization.

\begin{algorithm}[t]
\caption{\ourrl Training Process}
\label{alg:castflow}
\begin{algorithmic}[1]
\REQUIRE Reference policy $v^{\mathrm{ref}}$, training policy $v_\theta$ initialized from $v^{\mathrm{ref}}$;
\REQUIRE Reward model \reward, prompt set $\mathcal{C}$ with specified target texts;
\REQUIRE Group size $G$, optimized SDE steps $\mathcal{S}$, regional scale $\lambda$, regularization weight $\beta$.
\item[\textbf{Initialize:}] Data collection policy $v^{\mathrm{old}} \leftarrow v^{\mathrm{ref}}$.
\FOR{each training iteration}
    \STATE Sample a batch of prompts with target texts from $\mathcal{C}$
    \STATE \textcolor{gray}{// Rollout and defect-aware regional reward}
    \FOR{each prompt in the batch}
        \STATE Sample $G$ images $\{I_i\}_{i=1}^{G}$ with the SDE sampler (Eq.~\ref{eq:flowgrpo_sampler}), recording $n_k$ for $k\in\mathcal{S}$
        \STATE Predict regions and defect-aware transcriptions $\{(q_{i,o},t_{i,o})\}_{o=1}^{O_i}$ by \reward (Eq.~\ref{eq:textdart_prediction})
        \STATE Match regions to targets and compute regional rewards $r_{i,m}$ (Eq.~\ref{eq:regional_reward}) and global reward $R_i$ (Eq.~\ref{eq:global_text_reward})
    \ENDFOR
    \STATE \textcolor{gray}{// Target-group advantages.}
    \STATE Normalize rewards within each target group into $A_{i,m}$ (Eq.~\ref{eq:target_advantage}) and project onto the latent grid as $A_i^{\mathrm{reg}}(p)$ (Eq.~\ref{eq:advantage_projection})
    \STATE \textcolor{gray}{// Noise-aware Schedule.}
    \STATE Normalize $R_i$ within the prompt into $A_i^{\mathrm{glob}}$ and fuse $A_{i,k}(p)=\omega_k A_i^{\mathrm{glob}}+(1-\omega_k)A_i^{\mathrm{reg}}(p)$ (Eq.~\ref{eq:noise_aware_fusion})
    \STATE \textcolor{gray}{// Target-level policy ratios.}
    \FOR{each image $i$ and step $k\in\mathcal{S}$}
        \STATE Compute Guard-adjusted exponents $z_{i,k,p}$ and pool them into target ratios $\rho_{i,k,m}$ (Eq.~\ref{eq:target_policy_ratio})
    \ENDFOR
    \STATE Update $\theta$ by maximizing $\mathcal{J}_{\mathrm{SCANNER}}(\theta)$ (Eq.~\ref{eq:castflow_objective})
    \STATE $v^{\mathrm{old}} \leftarrow v_\theta$
\ENDFOR
\STATE \textbf{Output:} Optimized policy $v_\theta$
\end{algorithmic}
\end{algorithm}

\subsection{Experimental Setting Details}
\label{supp:exp-detail}

\begin{table}[t]
\centering
\caption{Hyperparameters for TextDART supervised fine-tuning.}
\label{tab:textdart_hyperparameters}
\small
\begin{tabular}{@{}ll@{}}
\toprule
Hyperparameter & Value \\
\midrule
Initialization & Qwen3-VL-8B-Instruct \\
Fine-tuning strategy & Full-parameter SFT \\
Trainable components & Language model, vision encoder, projector \\
GPUs / nodes & 32 / 4 \\
Training epochs & 4 \\
Batch size per GPU & 2 \\
Gradient accumulation steps & 2 \\
Effective batch size & 128 \\
Optimizer & AdamW (fused) \\
Learning rate & $5\times10^{-6}$ \\
Learning-rate schedule & Cosine \\
Warmup ratio & $0.05$ \\
Weight decay & $0.1$ \\
Adam $(\beta_1,\beta_2)$ & $(0.9,0.95)$ \\
Adam $\epsilon$ & $10^{-8}$ \\
Maximum gradient norm & $1.0$ \\
Maximum sequence length & 8,192 tokens \\
Image token budget & 2,304 \\
Overlength handling & Discard example \\
Precision & BF16 \\
Distributed training & DeepSpeed ZeRO-2 \\
Gradient checkpointing & Enabled \\
Attention implementation & FlashAttention \\
Random seed & 42 \\
\bottomrule
\end{tabular}
\end{table}

\begin{table}[t]
    \centering
    \caption{\textbf{Training and Evaluation Hyperparameters.} Settings for \reward with \ourrl.}
    \label{tab:training_hyperparameters}
    \footnotesize
    \setlength{\tabcolsep}{5pt}
    \renewcommand{\arraystretch}{1.05}
    \begin{tabular}{@{}lcc@{}}
        \toprule
        \textbf{Parameter} & \textbf{Qwen-Image-2512} & \textbf{FLUX.1-dev} \\
        \midrule
        \multicolumn{3}{l}{\textit{Model and optimization}} \\
        Training language & English / Chinese & English \\
        Resolution & $1024\times1024$ & $1024\times1024$ \\
        LoRA rank / alpha & $64$ / $128$ & $64$ / $128$ \\
        Mixed precision & BF16 & FP16 \\
        Optimizer & AdamW & AdamW \\
        Learning rate & $10^{-4}$ & $10^{-4}$ \\
        Weight decay & $10^{-4}$ & $10^{-4}$ \\
        Adam betas & $(0.9,0.999)$ & $(0.9,0.999)$ \\
        Adam epsilon & $10^{-8}$ & $10^{-8}$ \\
        Maximum gradient norm & $1.0$ & $1.0$ \\
        Per-device training batch size & $4$ & $3$ \\
        EMA decay & Disabled & $0.9$ \\
        Random seed & $42$ & $42$ \\
        \midrule
        \multicolumn{3}{l}{\textit{Sampling and evaluation}} \\
        Fast Flow-GRPO & Enabled & Enabled \\
        Rollout steps & $20$ & $28$ \\
        Evaluation steps & $50$ & $28$ \\
        Guidance scale (training / evaluation) & $4.0$ / $4.0$ & $3.5$ / $3.5$ \\
        Samples per prompt & $16$ & $24$ \\
        Flow-SDE noise level & $1.2$ & $0.9$ \\
        SDE window size & $5$ & $9$ \\
        \midrule
        \multicolumn{3}{l}{\textit{Reward and spatial credit assignment}} \\
        Defect mixing weight $w_d$ & $0.5$ & $0.5$ \\
        Defect-penalty scale $\kappa$ & $1.0$ & $1.0$ \\
        Regional scale $\lambda$ & $0.5$ & $0.5$ \\
        Target reward standard-deviation floor & $10^{-4}$ & $10^{-4}$ \\
        Advantage clipping bounds & $[-5,5]$ & $[-5,5]$ \\
        GRPO-Guard & Enabled & Enabled \\
        Policy clipping threshold & $10^{-4}$ & $10^{-4}$ \\
        Reference-penalty coefficient & $10^{-3}$ & $10^{-4}$ \\
        Reference weight (text / background) & $1$ / $2$ & $1$ / $2$ \\
        \bottomrule
    \end{tabular}
\end{table}

\subsubsection{Hyperparameters}
\paragraph{TextDART training.}
We initialize TextDART from Qwen3-VL-8B-Instruct~\citep{Qwen3-VL} and perform
full-parameter supervised fine-tuning, jointly updating the language
model, vision encoder, and multimodal projector.
Training uses 32 x H20 GPUs across four nodes.
Table~\ref{tab:textdart_hyperparameters} summarizes the settings.

\paragraph{Generative model post-training.}
We fine-tune both backbones with LoRA on 32 x H20 GPUs across four nodes, using DeepSpeed ZeRO-2 and gradient checkpointing. Table~\ref{tab:training_hyperparameters} lists all settings. Samples for the same prompt share initial noise, with independent perturbations during the selected Flow-SDE transitions. Qwen uses the official negative prompt during training and evaluation, while FLUX uses none. Our setup otherwise follows \citet{textpecker}.

\subsubsection{Details of Comparison with Baselines}
We reproduce the RL baselines with the open-source Flow-Factory framework~\citep{flowfactory}, starting from the same pretrained checkpoint for each backbone. All trained methods share the same training prompts, LoRA configuration, optimizer settings, resolution, sampling batch size, group size, and SDE window. We train Qwen-Image-2512 for 320 steps and FLUX.1-dev for 360 steps, requiring approximately 1,920 and 1,080 GPU hours, respectively.

\paragraph{Baseline Implementations.}
The base model is evaluated without RL fine-tuning. For OCRv5 with Flow-GRPO, we compute an image-level reward from the normalized edit distance between PP-OCRv5 transcriptions and the target text. For TextPecker with Flow-GRPO, we combine its semantic and defect scores with equal weights and a defect-penalty scale of $1$. Both RL baselines use Fast Flow-GRPO with GRPO-Guard, and reference regularization uses the backbone-specific coefficients in Table~\ref{tab:training_hyperparameters} with uniform spatial weighting.

\paragraph{Evaluation Protocol.}
All methods, including the pretrained model, are evaluated on the same benchmark prompts with the per-backbone evaluation settings in Table~\ref{tab:training_hyperparameters}. FLUX is evaluated in English and Qwen in English and Chinese, except on the English-only CVTG-2k. Each benchmark uses a common evaluator and metric configuration across methods, irrespective of the training reward. On \bench, we generate four images per prompt and set the evaluation defect-penalty scale to $5$.

\subsection{Extended Experiments}

\subsubsection{Evaluation of \reward}
\label{supp:exp-reward}
\begin{table*}[t]
\centering
\caption{\textbf{Evaluation of Text Recognizers on Canonical Text
Recognition (CTR), Text Structural Anomaly Perception
(TSAP), and Text Region Localization (TRL).} The best results are marked in \textbf{bold}. Results of non-supporting models are marked as $-$.}
\label{tab:reward_evaluation}
\scriptsize
\setlength{\tabcolsep}{3.2pt}
\renewcommand{\arraystretch}{1.12}
\resizebox{\textwidth}{!}{%
\begin{tabular}{l*{11}{c}}
\toprule
\multirow{2}{*}{Method}
& \multicolumn{2}{c}{CTR}
& \multicolumn{3}{c}{TSAP}
& \multicolumn{6}{c}{TRL} \\
\cmidrule(lr){2-3}\cmidrule(lr){4-6}\cmidrule(lr){7-12}
& R$\uparrow$ & NED$\downarrow$
& P$\uparrow$ & R$\uparrow$ & F1$\uparrow$
& P@.5$\uparrow$ & R@.5$\uparrow$ & F1@.5$\uparrow$
& P@.75$\uparrow$ & R@.75$\uparrow$ & F1@.75$\uparrow$ \\
\midrule
\multicolumn{12}{l}{\textbf{English}} \\
\addlinespace[3pt]
PP-OCRv5 & 0.7967 & 0.1191 & 0.0000 & 0.0000 & 0.0000 & 0.9070 & 0.9025 & 0.9048 & 0.6281 & 0.6250 & 0.6266 \\
MonkeyOCR & 0.8637 & 0.0788 & 0.0000 & 0.0000 & 0.0000 & 0.7262 & 0.5900 & 0.6510 & 0.4338 & 0.3525 & 0.3890 \\
GOT-OCR-2.0 & 0.6642 & 0.1903 & 0.0000 & 0.0000 & 0.0000 & $-$ & $-$ & $-$ & $-$ & $-$ & $-$ \\
Gemini-3.1-Pro-Preview & 0.8948 & 0.0523 & 0.5909 & 0.3588 & 0.4465 & 0.8433 & 0.8475 & 0.8454 & 0.4428 & 0.4450 & 0.4439 \\
GPT-5.5 & 0.8645 & 0.0669 & 0.5803 & 0.1680 & 0.2606 & 0.7487 & 0.7450 & 0.7469 & 0.4271 & 0.4250 & 0.4261 \\
TextPecker & 0.8943 & 0.0600 & 0.7509 & 0.4862 & 0.5902 & $-$ & $-$ & $-$ & $-$ & $-$ & $-$ \\
Qwen3-VL-8B & 0.8941 & 0.0576 & 0.0769 & 0.0005 & 0.0010 & 0.8650 & 0.8650 & 0.8650 & 0.5875 & 0.5875 & 0.5875 \\
\textbf{\reward (Ours)} & \textbf{0.9346} & \textbf{0.0362} & \textbf{0.7799} & \textbf{0.7874} & \textbf{0.7836} & \textbf{0.9750} & \textbf{0.9750} & \textbf{0.9750} & \textbf{0.9225} & \textbf{0.9225} & \textbf{0.9225} \\
\midrule
\multicolumn{12}{l}{\textbf{Chinese}} \\
\addlinespace[3pt]
PP-OCRv5 & 0.9530 & 0.0407 & 0.0000 & 0.0000 & 0.0000 & 0.9279 & 0.9217 & 0.9247 & 0.5638 & 0.5600 & 0.5619 \\
MonkeyOCR & 0.9363 & 0.0589 & 0.0000 & 0.0000 & 0.0000 & 0.7500 & 0.5650 & 0.6445 & 0.4624 & 0.3483 & 0.3973 \\
GOT-OCR-2.0 & 0.8939 & 0.0929 & 0.0000 & 0.0000 & 0.0000 & $-$ & $-$ & $-$ & $-$ & $-$ & $-$ \\
Gemini-3.1-Pro-Preview & 0.9621 & 0.0343 & 0.5841 & 0.3219 & 0.4150 & 0.8781 & 0.8883 & 0.8832 & 0.5568 & 0.5633 & 0.5601 \\
GPT-5.5 & 0.8115 & 0.1837 & 0.3626 & 0.1480 & 0.2102 & 0.7880 & 0.7683 & 0.7781 & 0.4342 & 0.4233 & 0.4287 \\
TextPecker & 0.8814 & 0.1141 & 0.7974 & 0.7123 & 0.7525 & $-$ & $-$ & $-$ & $-$ & $-$ & $-$ \\
Qwen3-VL-8B & 0.9502 & 0.0458 & 0.6000 & 0.0021 & 0.0041 & 0.9228 & 0.9167 & 0.9197 & 0.6946 & 0.6900 & 0.6923 \\
\textbf{\reward (Ours)} & \textbf{0.9639} & \textbf{0.0342} & \textbf{0.9033} & \textbf{0.7713} & \textbf{0.8321} & \textbf{0.9933} & \textbf{0.9917} & \textbf{0.9925} & \textbf{0.9399} & \textbf{0.9383} & \textbf{0.9391} \\
\bottomrule
\end{tabular}%
}
\end{table*}

% \paragraph{Tasks and Metrics.}
We evaluate \reward on the test set of \reward{}-123K (details in \ref{supp:textdart-123K}), where every method receives only the complete image. The first two tasks follow the evaluation protocol of \citet{textpecker}: \emph{Canonical Text Recognition (CTR)} measures the recovery of legible content through unit-level recall and normalized edit distance over all images, including defective ones, while \emph{Text Structural Anomaly Perception (TSAP)} measures whether malformed text is explicitly identified, using micro precision, recall, and $F_1$ over counted anomaly markers, with all methods rescored under this common protocol. We additionally introduce \emph{Text Region Localization (TRL)}, which compares the union of predicted and ground-truth polygons at IoU thresholds of $0.5$ and $0.75$, accommodating different box granularities.
We compare against specialist OCR systems~\citep{ppocrv5,got-ocrv2,monkeyocr}, the structure-aware recognizer~\citep{textpecker} and multimodal models~\citep{Qwen3-VL,gpt5p5,gemini3p1propreview}. 
GOT-OCR-2.0 and TextPecker produce only transcriptions, so their localization entries are unavailable.
% \paragraph{Quantitative Assessment.}
\reward achieves the best results on every metric in Tab.~\ref{tab:reward_evaluation}. The CTR gap over the strongest proprietary models is moderate, as general models already read clean text well. The decisive gap lies in defect perception: OCR systems score zero on TSAP, as they cannot report malformed glyphs at all. Even the best baseline TextPecker trails by $0.19$ and $0.08$ $F_1$ in English and Chinese, and the unfine-tuned backbone scores nearly zero, showing that defect awareness comes from \reward-123K fine-tuning rather than the backbone itself. The localization margin also widens at the stricter IoU $0.75$, indicating precise boundaries rather than coarse coverage. These results validate \reward as a reliable reward source for region-level credit assignment.

\subsubsection{User Study}
\label{app:user_study}

We conduct a blind randomized user study with 10 participants to complement the automatic evaluation with human judgments.
For each sample, outputs from \reward + \ourrl, TextPecker + Flow-GRPO, OCRv5 + Flow-GRPO, and the Qwen-Image-2512 base model are anonymized and presented in a randomly shuffled order.
Participants are unaware of the model identities and independently select the best result according to two criteria: \textbf{text accuracy}, which evaluates whether the rendered text is correct and legible, and \textbf{text aesthetics}, which evaluates the visual quality and naturalness of the rendered text.
We report the average win rate aggregated over all samples and annotators.

As shown in Table~\ref{tab:user_study}, \reward + \ourrl receives the highest human preference on both dimensions, achieving a win rate of \textbf{46.8\%} for text accuracy and \textbf{39.4\%} for text aesthetics.
Notably, OCRv5 + Flow-GRPO receives substantially lower human preference despite explicitly optimizing recognition-based rewards, suggesting that improved recognizer scores do not necessarily translate into better perceived text rendering quality.

\begin{table}[ht!]
    \centering
    \caption{
    \footnotesize
    \textbf{Blind Randomized User Study on Visual Text Generation.}
    We report the average win rate across all samples and annotators.
    The best result in each row is highlighted in bold.
    }
    \label{tab:user_study}
    \scriptsize
    \setlength{\tabcolsep}{4pt}
    \begin{tabular}{@{}lcccc@{}}
        \toprule
        \textbf{Metric} &
        \shortstack{Qwen-Image-2512} & 
        \shortstack{OCRv5 +\\Flow-GRPO} &
        \shortstack{TextPecker +\\Flow-GRPO} &
        \shortstack{\reward +\\\ourrl} \\
        \midrule
        Text Accuracy $\uparrow$
            & 21.30\% & 9.40\% & 22.50\% & \textbf{46.80\%}  \\
        Text Aesthetics $\uparrow$
            & 24.80\% & 13.10\% & 22.70\% & \textbf{39.40\%} \\
        \bottomrule
    \end{tabular}
\end{table}

\subsubsection{Additional Ablation Study}
\label{supp:ablation}
\paragraph{Spatially Weighted KL Regularization.}
% visual【ok，flux，hack】KL
% global KL 与不同 box/background 权重
\begin{figure*}[!t]
% \vspace{-10pt}
  \centering
  % \begin{subfigure}{\linewidth}
    % 将这里的 width=\linewidth 改为一个更小的值，比如 0.8\linewidth
    \includegraphics[width=1.0\textwidth]{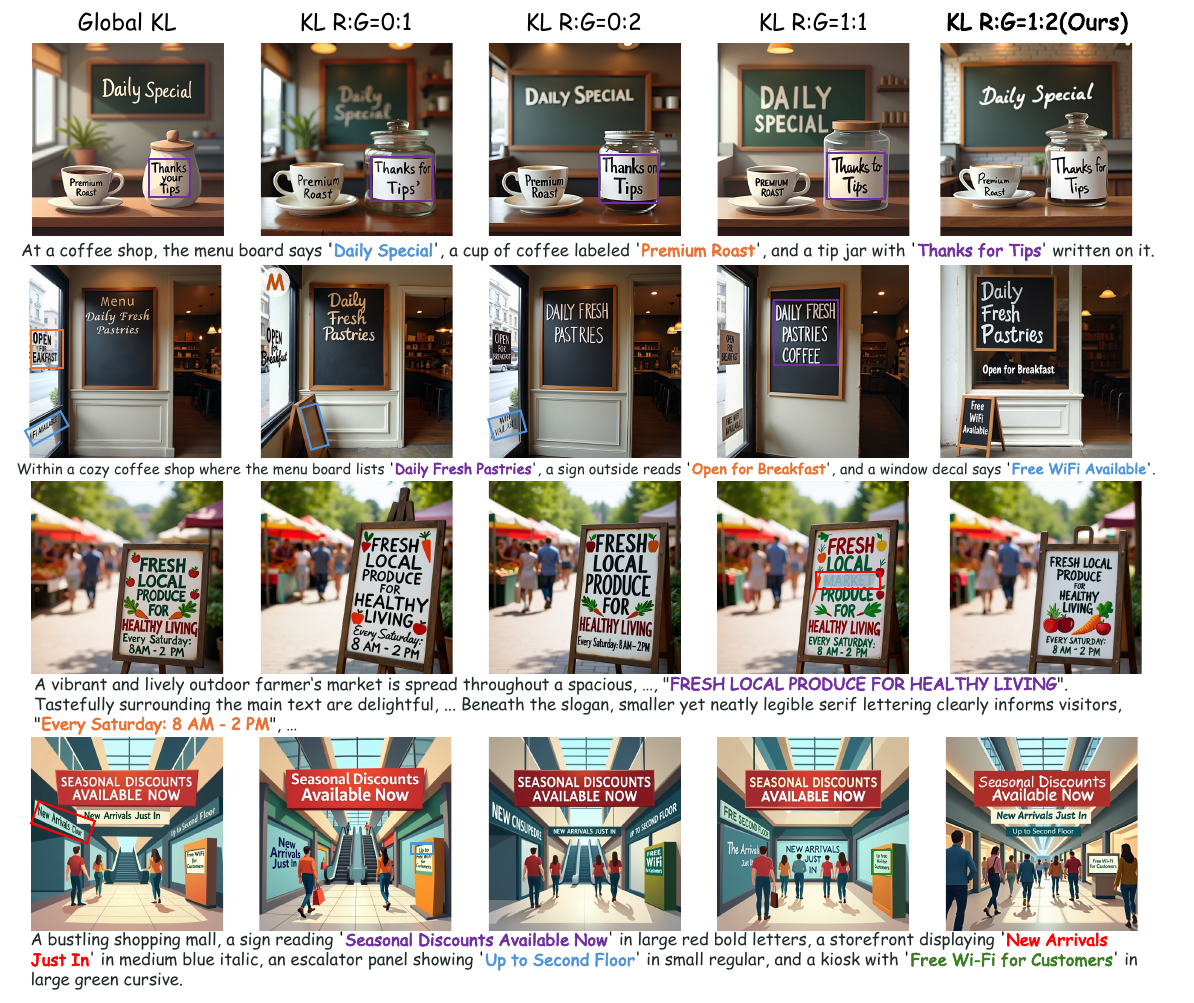}
    \vspace{-20pt}
    \caption{\textbf{Ablation on Spatially Weighted KL Regularization.} Global KL leaves residual misspellings, text-free penalties cause hallucinated text, and equal weights induce reward hacking (covering text with plain blocks, row 3). Our background-weighted variant (R:G $=$ 1:2) renders all target text correctly while preserving the scene.}
    \label{fig: ablation-kl}
  % \end{subfigure}
\end{figure*}
Fig.~\ref{fig: ablation-kl} compares global KL regularization with spatially weighted variants of different text-region (R) and background (G) weights. Global KL applies the penalty uniformly, which counteracts the text reward on text tokens and leaves residual misspellings (rows 1 and 4). Removing the penalty on text regions (R:G $=$ 0:1 and 0:2) frees text optimization but weakens the anchor on content, leading to hallucinated text such as the extra letter and the garbled sign in the second row. With equal weights (R:G $=$ 1:1), reward hacking emerges, where the model covers text regions with plain blocks to escape defect penalties (third row). Our choice (R:G $=$ 1:2) keeps text regions free to improve while anchoring the background more strongly, yielding correct text and faithful scenes without hacking behavior.

% ===========================
% ==========Figure============
% ===========================

\begin{figure*}[!t]
% \vspace{-10pt}
  \centering
  % \begin{subfigure}{\linewidth}
    % 将这里的 width=\linewidth 改为一个更小的值，比如 0.8\linewidth
    \includegraphics[width=1.0\textwidth]{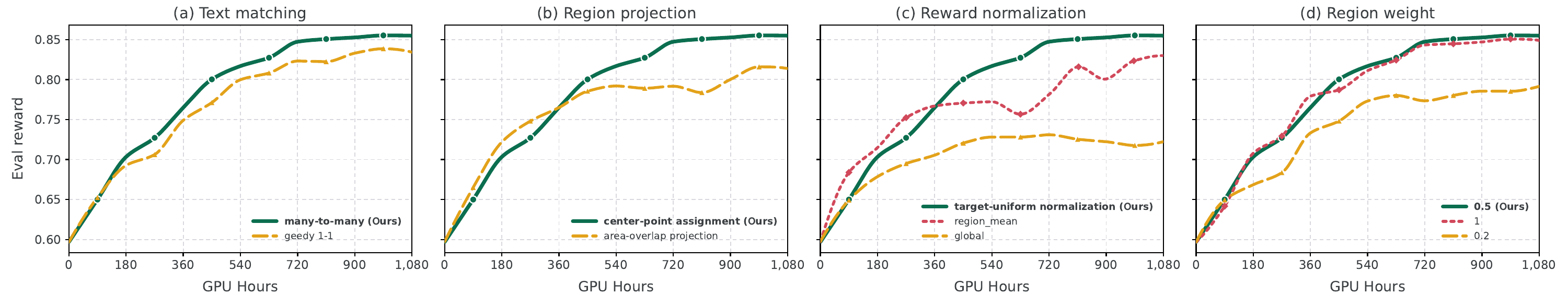}
    % \vspace{-20pt}
    \caption{\textbf{Supplementary Ablations.} Training curves of evaluation reward for (a) region--target matching, (b) image-to-latent projection, (c) normalization of regional contributions, and (d) the regional scale $\lambda$. Our choices (green) converge the highest in all four comparisons.}
    \label{fig: curve-supp}
  % \end{subfigure}
\end{figure*}

\paragraph{Region--Target Matching.}
Fig.~\ref{fig: curve-supp}(a) compares region--target matching strategies. Greedy one-to-one matching plateaus well below ours, as it mismatches regions and targets whenever one region covers multiple targets or one target spans several regions, corrupting the regional rewards. Many-to-many matching with merging preserves the correct correspondence and reaches a consistently higher reward.

\paragraph{Projection of Regional Advantages onto Latent Tokens.}
Fig.~\ref{fig: curve-supp}(b) compares image-to-latent projection rules. Area-overlap projection rises slightly faster at first but plateaus with oscillation, as partially covered tokens at region boundaries receive advantages from neighboring content. Assigning each token by its center point keeps region boundaries clean and converges markedly higher.

\paragraph{Area Normalization of Regional Contributions.}
Fig.~\ref{fig: curve-supp}(c) compares normalization schemes for regional contributions. Without normalization (global), larger regions dominate the updates and small text remains under-corrected, yielding the lowest curve. Normalizing per detected region (region-mean) improves over it but stays unstable, since the weighting depends on how text is partitioned into regions rather than on semantic targets. Normalizing each merged target's contribution by its support size (ours) is both stable and the highest.

\paragraph{Sensitivity to Regional Loss Weight.}
Fig.~\ref{fig: curve-supp}(d) varies the regional scale $\lambda$. Reducing it to $0.2$ clearly slows learning, as regional signals become too weak against the global branch, while $0.5$ and $1.0$ behave almost identically with a slight edge for $0.5$. Our method is thus robust to $\lambda$ as long as regional credit is not under-weighted.

\subsubsection{Effect of Replacing the Text Recognizer}
\begin{figure*}[!t]
% \vspace{-10pt}
  \centering
  % \begin{subfigure}{\linewidth}
    % 将这里的 width=\linewidth 改为一个更小的值，比如 0.8\linewidth
    \includegraphics[width=0.8\textwidth]{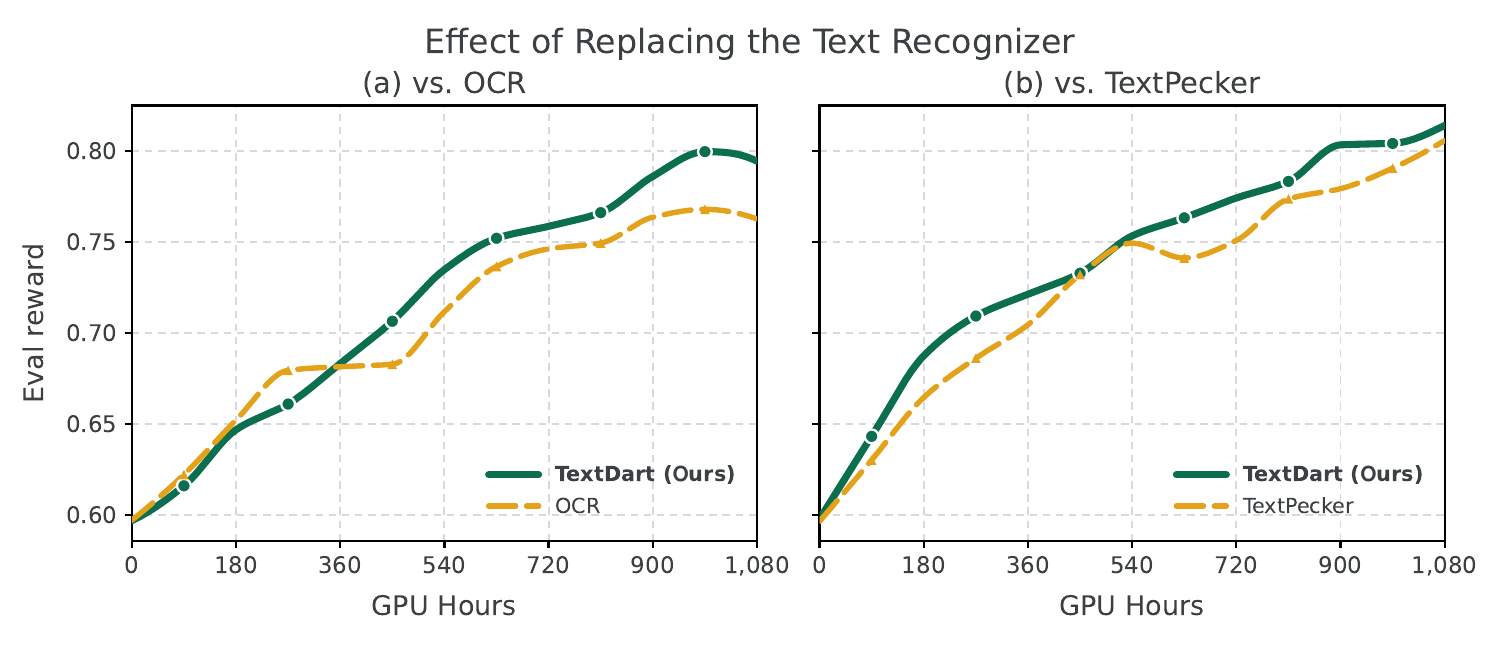}
    \vspace{-15pt}
    \caption{\textbf{Effect of Replacing the Text Recognizer.} Training curves with the downstream pipeline fixed. (a) The OCR reward saturates and declines late in training, while \reward keeps improving. (b) \reward also surpasses TextPecker with a steadier curve, showing the benefit of region-precise feedback.}
    \label{fig: replace}
  % \end{subfigure}
\end{figure*}
% OCR识别：对比ocr换成\reward效果，和对比textpecker换成\reward效果
% 只show flux的
% 后续reward计算和flow-grpo过程完全一致

Fig.~\ref{fig: replace} isolates the contribution of the text recognizer by replacing OCRv5 or TextPecker with \reward, while keeping downstream reward computation and Flow-GRPO optimization unchanged. Against OCR (Fig.~\ref{fig: replace}a), both variants learn similarly at first, but the OCR-based variant saturates and even declines late in training, as it misreads malformed glyphs and stops providing useful signal once most text is nearly correct, whereas \reward keeps improving through defect-aware feedback. Against TextPecker (Fig.~\ref{fig: replace}b), the gap is smaller since glyph defects are already penalized, yet \reward remains consistently higher with a steadier curve, indicating that region-precise feedback optimizes more effectively than whole-image defect scoring.

\subsubsection{\ourrl Generalization across Reward Models}
\begin{table*}[t]
\centering
\caption{
\textbf{Generalization of \ourrl across Reward Models.}
Both adaptations follow the same region-wise reward computation
paradigm as \reward and the same subsequent \ourrl
training procedure.
The best results are marked in \textbf{bold}.
Results for unsupported languages are marked as ``$-$''.}
\label{tab:reward_generalization}
\small
\setlength{\tabcolsep}{3pt}
\renewcommand{\arraystretch}{1.12}
\begin{adjustbox}{max width=\textwidth}
\begin{tabular}{@{}ll*{12}{c}@{}}
\toprule
\multirow{2}{*}{Reward Model}
& \multirow{2}{*}{RL Algorithm}
& \shortstack{LongText- Bench}
& \multicolumn{4}{c}{OneIG-Bench}
& \multicolumn{2}{c}{CVTG-2k}
& \multicolumn{2}{c}{GenTextEval-Bench}
& \multicolumn{3}{c}{\bench} \\
\cmidrule(lr){3-3}
\cmidrule(lr){4-7}
\cmidrule(lr){8-9}
\cmidrule(lr){10-11}
\cmidrule(lr){12-14}
& & \shortstack{TextScore $\uparrow$}
& \shortstack{TextScore $\uparrow$}
& ED $\downarrow$ & CR $\uparrow$ & WAC $\uparrow$
& \shortstack{WordAcc $\uparrow$}
& NED $\uparrow$ & Qua $\uparrow$ & Sem $\uparrow$ & CS $\uparrow$ & CG $\uparrow$ & OG $\uparrow$ \\
\midrule
\multirow{2}{*}{OCRv5} & Flow-GRPO
& 0.769 & 0.751 & 64.55 & 0.220 & 0.506
& 0.686 & 0.818 & 0.838 & 0.589 & 0.703 & 0.726 & 0.365 \\
& \ourrl
& 0.809 & 0.771 & 66.29 & 0.230 & \textbf{0.552}
& \textbf{0.809} & \textbf{0.905} & \textbf{0.862} & 0.591 & 0.797 & 0.840 & 0.368 \\
\midrule
\multirow{2}{*}{TextPecker} & Flow-GRPO
& 0.801 & 0.770 & 63.19 & 0.221 & 0.533
& 0.772 & 0.879 & 0.836 & 0.582 & 0.789 & 0.809 & 0.405 \\
& \ourrl
& \textbf{0.813} & \textbf{0.804} & \textbf{58.10}
& \textbf{0.253} & 0.548
& 0.807 & 0.901
& 0.840 & \textbf{0.593} & \textbf{0.864} & \textbf{0.897} & \textbf{0.583} \\
\bottomrule
\end{tabular}
\end{adjustbox}
\end{table*}

Tab.~\ref{tab:reward_generalization} applies \ourrl to existing reward models.
Both adaptations first localize text regions with PP-OCRv5 and then follow the
same region-wise reward computation and \ourrl training procedure as \reward.
OCRv5 directly transcribes each detected region. TextPecker does not localize
text from a full image; instead, we provide the full image together with each
detected box and query it separately to obtain defect-aware transcriptions.
Since each image incurs one request per detected box, this adaptation introduces
additional inference overhead and is used only for this generalization study.
\ourrl improves Flow-GRPO on most metrics under both rewards, with large gains
on text-specific metrics such as WordAcc from $0.686$ to $0.809$ with OCRv5 and
ED from $63.19$ to $58.10$ with TextPecker, showing that spatially aligned
credit assignment is largely orthogonal to the choice of reward model. Two
further observations stand out. First, the gains are generally larger with the
defect-aware TextPecker than with OCRv5, whose ED slightly degrades, as regional
credit can only amplify what the reward perceives and a recognition-only reward
remains blind to glyph defects. Second, both adapted variants still fall short
of the full combination of \reward and \ourrl in Tab.~\ref{tab:main_results},
confirming that reward quality and credit assignment are complementary rather
than interchangeable.

\subsubsection{Multi-Reward Optimization}

\begin{figure*}[!t]
% \vspace{-10pt}
  \centering
  % \begin{subfigure}{\linewidth}
    % 将这里的 width=\linewidth 改为一个更小的值，比如 0.8\linewidth
    \includegraphics[width=1.0\textwidth]{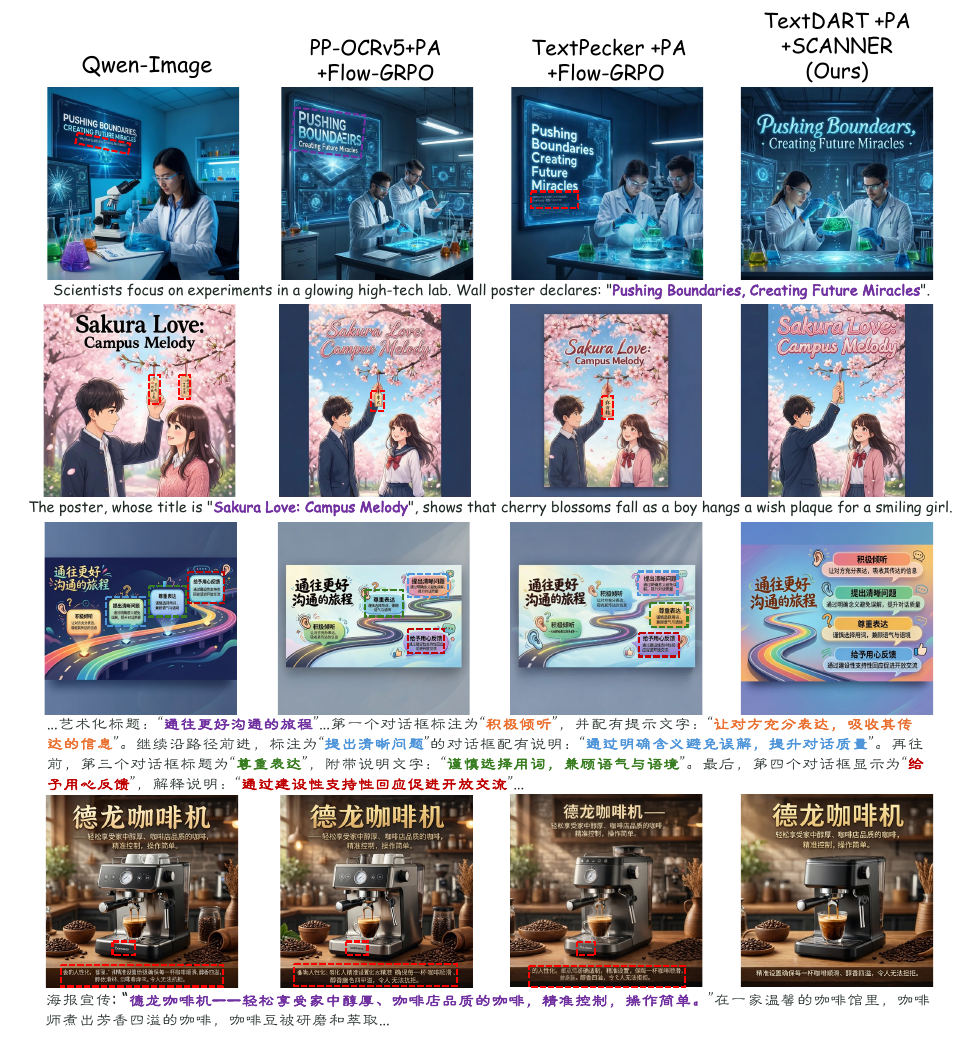}
    \vspace{-20pt}
    \caption{\textbf{Qualitative Results of Multi-reward Optimization on Qwen-Image-2512.} Each text reward is combined with the PickScore\citep{pickscore} and aesthetic\citep{aes} rewards (PA). Red dashed boxes mark text errors. \reward with \ourrl renders all target text correctly while producing the most appealing layout, color, and composition, whereas the baselines garble text or degrade the scene.}
    \label{fig: multi-reward}
  % \end{subfigure}
\end{figure*}

\begin{table*}[t]
\centering
\caption{
\textbf{Multi-reward Optimization.}
Each text reward (OCRv5, TextPecker, or \reward) is combined with the PickScore~\citep{pickscore} and aesthetic~\citep{aes} rewards (PA) under its respective RL algorithm, and Base denotes the model without RL fine-tuning.
Combining \reward with PA under \ourrl achieves the best results on all text
metrics in both languages while maintaining good visual appeal
(Fig.~\ref{fig: multi-reward}), showing that superior text quality is attained
together with, rather than at the cost of, general image quality.
The best results within each language group are marked in \textbf{bold}, and $-$ marks unsupported languages.}
\label{tab:multi-reward}
\small
\setlength{\tabcolsep}{3pt}
\renewcommand{\arraystretch}{1.12}
\begin{adjustbox}{max width=\textwidth}
\begin{tabular}{@{}l*{12}{c}@{}}
\toprule
\multirow{2}{*}{Variant}
& \shortstack{LongText-Bench}
& \multicolumn{4}{c}{OneIG-Bench}
& \multicolumn{2}{c}{CVTG-2k}
& \multicolumn{2}{c}{GenTextEval-Bench}
& \multicolumn{3}{c}{\bench} \\
\cmidrule(lr){2-2}
\cmidrule(lr){3-6}
\cmidrule(lr){7-8}
\cmidrule(lr){9-10}
\cmidrule(lr){11-13}
& \shortstack{TextScore $\uparrow$}
& \shortstack{TextScore $\uparrow$}
& ED $\downarrow$ & CR $\uparrow$ & WAC $\uparrow$
& \shortstack{WordAcc $\uparrow$}
& NED $\uparrow$ & Qua $\uparrow$ & Sem $\uparrow$ & CS $\uparrow$ & CG $\uparrow$ & OG $\uparrow$ \\
\midrule
\multicolumn{13}{l}{\textit{English Text Generation}} \\
\addlinespace[2pt]
Base & 0.952 & 0.986 & 24.54 & 0.461 & 0.895 & 0.844 & 0.922 & 0.944 & 0.787 & 0.887 & 0.899 & 0.252 \\
OCRv5 + PA / Flow-GRPO & 0.951 & 0.992 & 18.80 & 0.507 & 0.906 & 0.848 & 0.924 & 0.950 & 0.796 & 0.880 & 0.889 & 0.262 \\
TextPecker + PA / Flow-GRPO & 0.957 & 0.992 & 17.21 & 0.505 & 0.903 & 0.849 & 0.925 & 0.945 & 0.796 & 0.880 & 0.883 & 0.212 \\
\reward + PA / \ourrl & \textbf{0.969} & \textbf{0.994} & \textbf{14.65} & \textbf{0.571} & \textbf{0.907} & \textbf{0.880} & \textbf{0.940} & \textbf{0.982} & \textbf{0.855} & \textbf{0.957} & \textbf{0.976} & \textbf{0.434} \\
\midrule
\multicolumn{13}{l}{\textit{Chinese Text Generation}} \\
\addlinespace[2pt]
Base & 0.966 & 0.983 & 13.06 & 0.554 & 0.851 & $-$ & $-$ & 0.955 & 0.853 & 0.931 & 0.945 & 0.198 \\
OCRv5 + PA / Flow-GRPO & 0.967 & 0.986 & 11.27 & 0.584 & 0.851 & $-$ & $-$ & 0.956 & 0.855 & 0.935 & 0.936 & 0.209 \\
TextPecker + PA / Flow-GRPO & 0.969 & 0.987 & 10.85 & 0.571 & 0.856 & $-$ & $-$ & 0.960 & 0.870 & 0.935 & 0.943 & 0.177 \\
\reward + PA / \ourrl & \textbf{0.983} & \textbf{0.993} & \textbf{8.21} & \textbf{0.706} & \textbf{0.857} & $-$ & $-$ & \textbf{0.975} & \textbf{0.897} & \textbf{0.958} & \textbf{0.973} & \textbf{0.414} \\
\bottomrule
\end{tabular}
\end{adjustbox}
\end{table*}

We further investigate whether \reward can be combined with general-purpose rewards under \ourrl to jointly optimize textual fidelity and overall image quality. Following ~\cite{gdpo}, we normalize each reward within its prompt group and sum the resulting advantages. This integrates naturally with \ourrl: the PickScore and aesthetic rewards (PA) are image-level, so their advantages are uniform over all latent tokens, while the text reward contributes its spatially aligned advantage map, and their sum enters the same target-level policy objective unchanged. We combine each text reward (OCRv5, TextPecker, or \reward) with PA and perform RL fine-tuning.

As shown in Tab.~\ref{tab:multi-reward}, \reward{} + PA with \ourrl achieves the strongest results on all metrics in both English and Chinese, outperforming OCRv5 + PA and TextPecker + PA under Flow-GRPO. Qualitative results in Fig.~\ref{fig: multi-reward} further show that this combination renders text correctly while preserving visual quality and composition, indicating that \reward with \ourrl serves as a flexible component in multi-reward optimization.

\subsection{Additional Qualitative Results for Localization and Defect-Aware Recognition}
\begin{figure}[!t]
% \vspace{-10pt}
  \centering
    \includegraphics[width=0.85\textwidth]{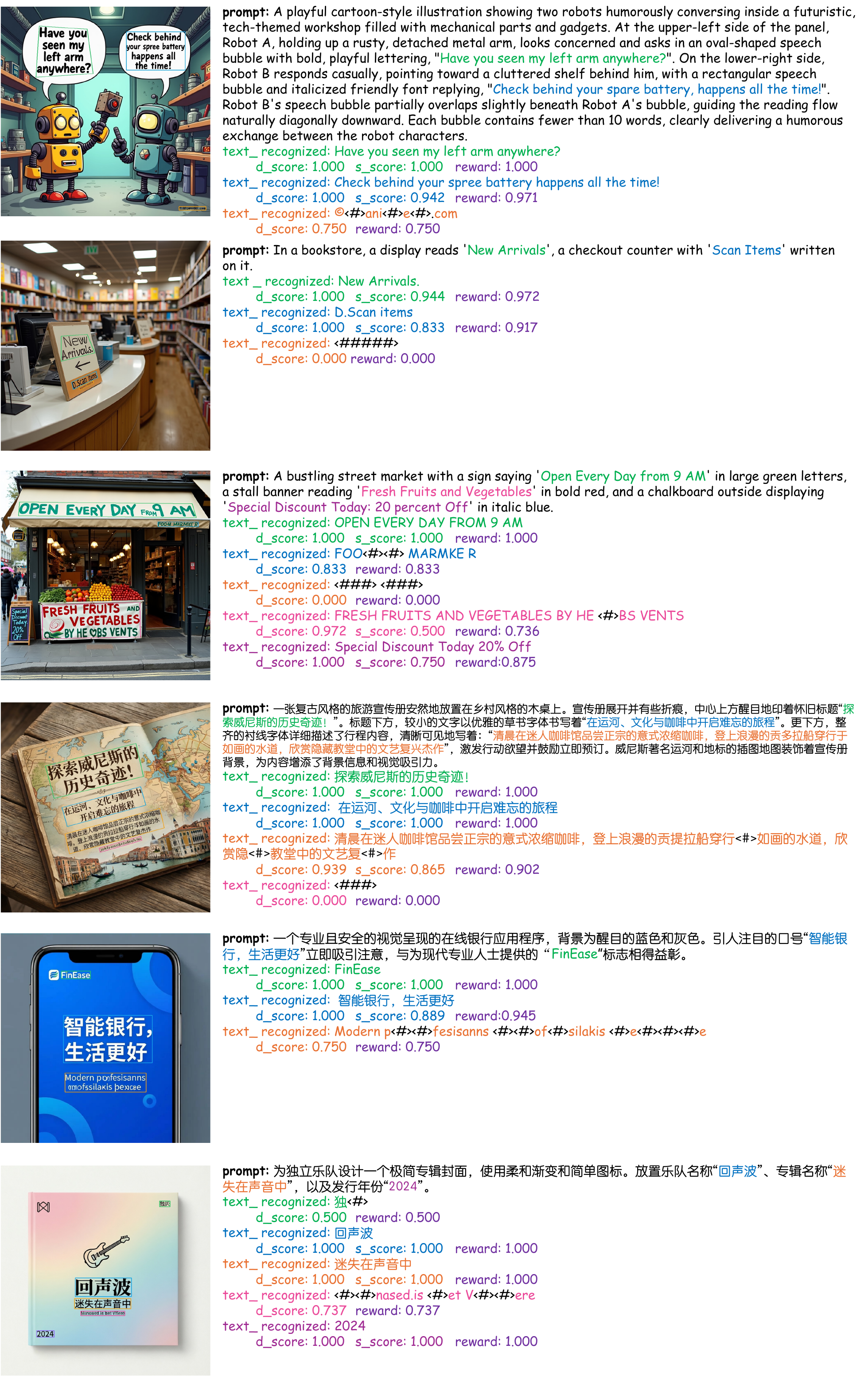}
    % \vspace{-10pt}
    \caption{\textbf{Localization and Recognition Results of \reward.} Each detected region is annotated with its defect-aware transcription and glyph-quality score (d\_score), and regions matched to a target additionally receive a semantic-correctness score (s\_score).}
    \label{fig: vis-reward-more}
\end{figure}

Fig.~\ref{fig: vis-reward-more} presents more cases of \reward's region-level recognition, where each detected region is transcribed with defect markers and scored for glyph quality and semantic correctness.

\subsection{Additional Qualitative Results for Visual Text Generation}

\begin{figure}[!t]
\vspace{-20pt}
  \centering
    \includegraphics[width=0.8\textwidth]{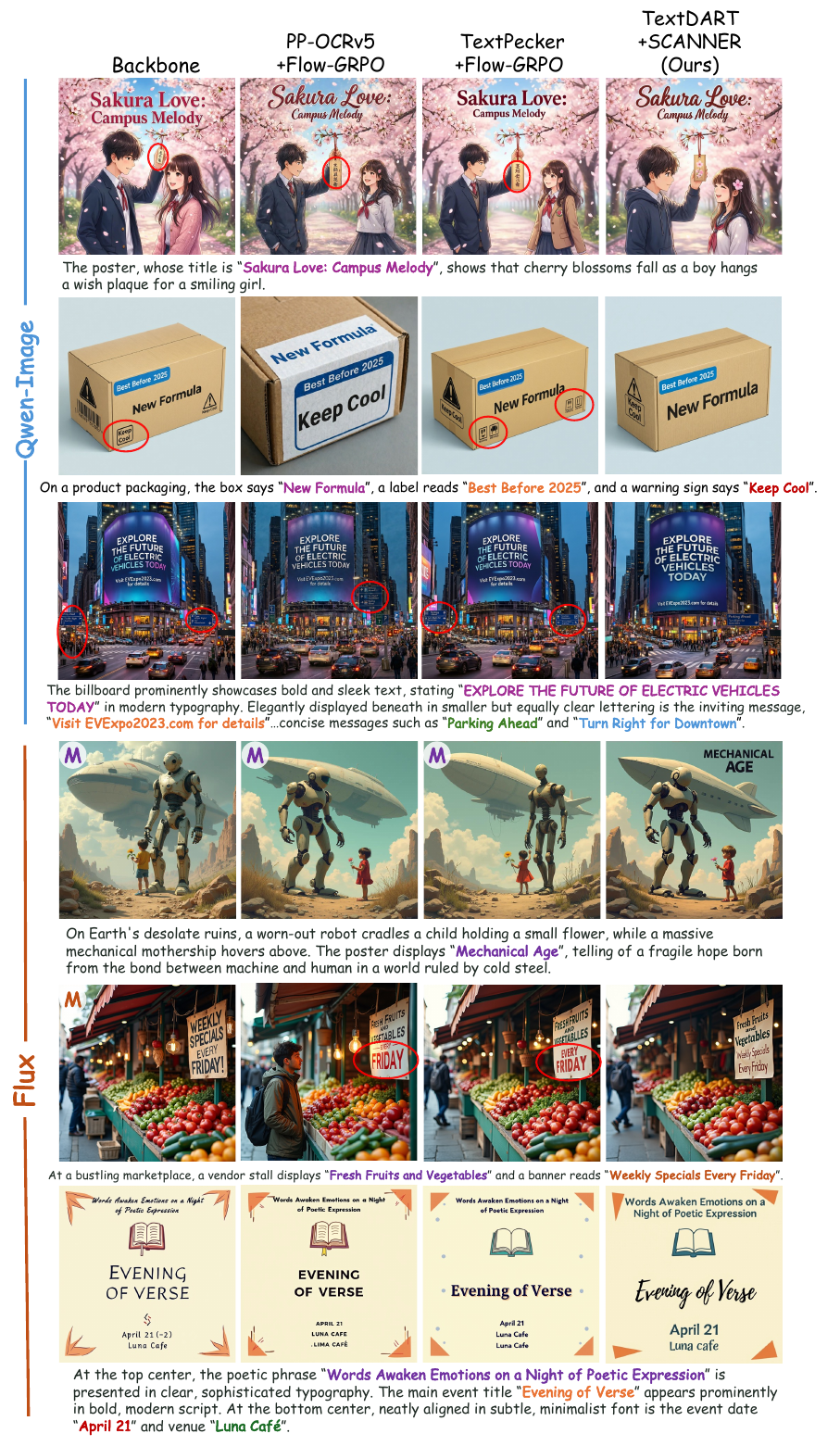}
    \caption{\textbf{Qualitative Comparison on English Prompts.} Rows 1--3 are generated on Qwen-Image and rows 4--6 on Flux, and circles highlight text errors of the baselines.}
    \label{fig: vis-supp-en}
    \vspace{-20pt}
\end{figure}

\paragraph{Results on English Prompts.}
Fig.~\ref{fig: vis-supp-en} presents additional English results on Qwen-Image (rows 1--3) and Flux (rows 4--6), where the trends observed in the main text hold throughout. Baseline errors concentrate on small secondary text, such as the wish plaque in row 1, the warning sign in row 2, and the street signs in row 3, where unrequested regions are filled with garbled strokes. The baselines also miss or duplicate content, rendering only a single letter as the poster title in row 4, producing incomplete signage in row 5, and repeating the venue line in row 6. Our method renders every requested target correctly across all rows while keeping spontaneous text free of garbled writing.

\begin{figure}[!t]
\vspace{-20pt}
  \centering
    \includegraphics[width=0.7\textwidth]{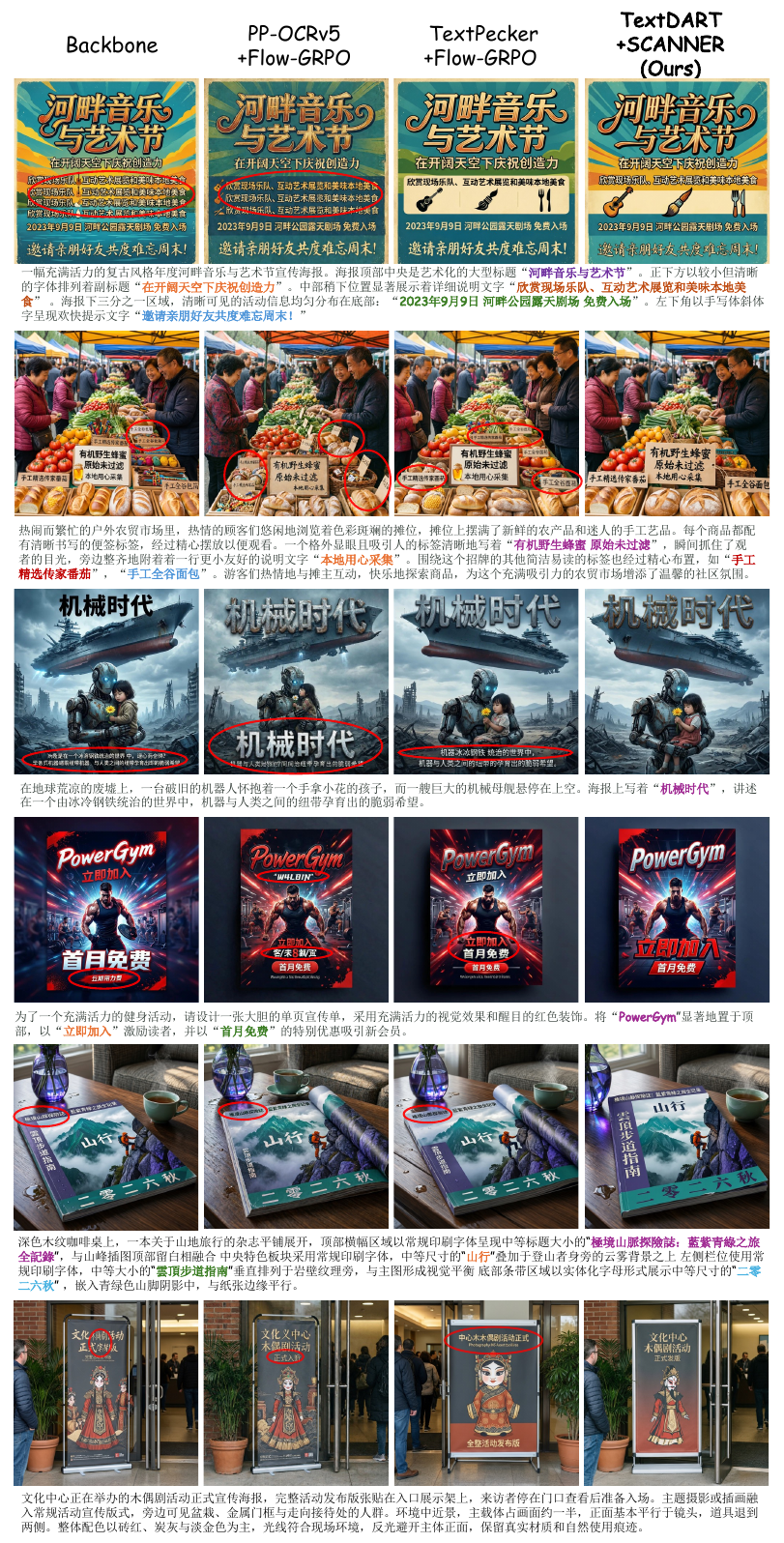}
    \caption{\textbf{Qualitative Comparison on Chinese Prompts.} All images are generated on Qwen-Image, and circles highlight text errors of the baselines.}
    \label{fig: vis-supp-zh}
    \vspace{-20pt}
\end{figure}

\paragraph{Results on Chinese Prompts.}
Fig.~\ref{fig: vis-supp-zh} presents additional Chinese results on Qwen-Image. The backbone and the OCRv5 reward repeat the middle descriptive line several times in row 1, and all baselines garble the small price tags in row 2 and the long top banner in row 5. Rows 3 and 4 exhibit the same failure on spontaneous and secondary text, where the baselines append garbled small print or corrupt the call-to-action slogans, and row 6 contains wrong or repeated characters in the poster header. Our method renders every requested target correctly, from multi-line posters to dense magazine spreads, and produces no garbled extra text.

\subsection{Additional Details on \bench}
\label{supp:bench}

\paragraph{Comparison with Existing Benchmarks.}
Table~\ref{tab:benchmark_comparison} compares \bench with existing visual-text benchmarks. While multi-block prompts and, in most cases, both English and Chinese are already covered, glyph defects beyond transcription accuracy are scored only by GenTextEval, and none of them evaluates text that may emerge without being prescribed. \bench is the only benchmark that covers all four aspects.

\begin{table*}[t]
\centering
\caption{\textbf{Comparison of Visual Text Generation Benchmarks.}
The comparison concerns the released prompt sets and their original
text-evaluation protocols.
Multi-block prompts specify multiple spatially separated text blocks.
Glyph-defect scoring explicitly assesses malformed glyphs beyond
transcription accuracy.
Text-optional evaluation assesses text that may emerge without
prescribing its presence or content.
For OneIG-Bench, we consider its text-rendering subset.
\checkmark: included; $\times$: not included.}
\label{tab:benchmark_comparison}
\scriptsize
\setlength{\tabcolsep}{4pt}
\begin{tabular}{lcccc}
\toprule
Benchmark
& \shortstack{Multi-block\\prompts}
& \shortstack{English +\\Chinese}
& \shortstack{Glyph-defect\\scoring}
& \shortstack{Text-optional\\evaluation} \\
\midrule
LongText-Bench~\citep{longtextbench}
& \checkmark & \checkmark & $\times$ & $\times$ \\
OneIG-Bench~\citep{oneig}
& \checkmark & \checkmark & $\times$ & $\times$ \\
CVTG-2K~\citep{cvtg2k}
& \checkmark & $\times$ & $\times$ & $\times$ \\
GenTextEval~\citep{textpecker}
& \checkmark & \checkmark & \checkmark & $\times$ \\
\midrule
\bench{} (ours)
& \checkmark & \checkmark & \checkmark & \checkmark \\
\bottomrule
\end{tabular}
\end{table*}

\begin{table*}[t]
\centering
\caption{\textbf{Construction Branches (left) and Image Sources (right) of the \reward-123K Training Split.} Percentages use all 122,200 training entries. The graphic design collections category covers existing text-rich design images, including posters, logos, infographics, and artistic lettering.}
\label{tab:textdart-sources}
\small
\begin{minipage}[t]{0.52\textwidth}
\centering
\setlength{\tabcolsep}{5pt}
\begin{tabular}{lrr}
\toprule
Branch & Entries & Share \\
\midrule
Controlled synthesis & 18,856 & 15.4\% \\
Caption regeneration & 30,423 & 24.9\% \\
Graphic design collections & 26,915 & 22.0\% \\
Long-tail & 13,885 & 11.4\% \\
Small-text & 10,273 & 8.4\% \\
Text-free & 21,848 & 17.9\% \\
\midrule
Total & 122,200 & 100.0\% \\
\bottomrule
\end{tabular}
\end{minipage}
\hfill
\begin{minipage}[t]{0.42\textwidth}
\centering
\setlength{\tabcolsep}{5pt}
\begin{tabular}{lrr}
\toprule
Source & Entries & Share \\
\midrule
Open-source models & 47,828 & 39.1\% \\
Closed-source models & 49,284 & 40.4\% \\
Real-world imagery & 25,088 & 20.5\% \\
\midrule
Total & 122,200 & 100.0\% \\
\bottomrule
\end{tabular}
\end{minipage}
\end{table*}

\begin{table}[t]
\centering
\caption{\textbf{Construction Dimensions of the Controlled Prompts.} Distributions are recorded on the specifications of the controlled subset and listed in the order of the corresponding categories. $^{*}$Single-line vs.\ multi-line regions. $^{\dagger}$Recorded over the first three categories.}
\label{tab:textdart-dimensions}
\small
\setlength{\tabcolsep}{4pt}
\begin{tabular}{@{}p{0.14\columnwidth}p{0.55\columnwidth}p{0.21\columnwidth}@{}}
\toprule
Dimension & Categories & Distribution (\%) \\
\midrule
Scene & Street views, posters, infographics, etc.\ (17 in total) & 5.8--5.9 each \\
Script & Chinese, English, mixed, digits and symbols & -- \\
Text content & Frequency-balanced characters and lexical units, spanning short labels to long phrases & -- \\
Composition & Single, multiple, and dense regions, 1--10 lines each & 49.3 / 50.7$^{*}$ \\
Geometry & Horizontal, vertical, rotated, perspective & 42.4 / 21.7 / 21.6 / 14.3 \\
Typography & Standard, handwriting or calligraphy, bold, artistic lettering & 39.9 / 40.0 / 20.1$^{\dagger}$ \\
Scale & Large display, medium, small & 30.1 / 39.9 / 30.0 \\
Text presence & Text-containing, text-free & -- \\
\bottomrule
\end{tabular}
\end{table}

\paragraph{Prompt Construction.}
We build prompts from structured scene specifications that combine a physical setting, a text-bearing object, a theme, a composition, and a color palette, and expand them with a language model~\citep{guo2026less} into full descriptions covering material, artwork, props, background, and lighting, so that text appears on functional objects in concrete scenes rather than as isolated lettering. Critical prompts additionally specify one to five target strings with roles such as titles, menu entries, product labels, and headings, together with their described locations and visual treatments. Optional prompts instead describe the object's ordinary real-world function, such as a published book offered for sale or a menu consulted by a diner, without prescribing any wording, so that text is a natural possibility rather than an explicit instruction.

\paragraph{Dataset Statistics.}
\bench contains 600 prompts organized as 300 critical--optional pairs balanced between English and Chinese, where each pair shares a scene category and theme and differs in whether text is prescribed. The pairs span ten scene categories, including book and magazine covers, posters, menus, product packaging, storefronts, tickets and programs, information boards, screen displays, and album and game covers, with category sizes ranging from 10 to 44 pairs and both languages represented in every category. The 300 critical prompts contain 783 target strings (404 English and 379 Chinese, 2.6 per prompt on average), with lengths of 3--76 characters for English and 1--19 for Chinese, counting literal characters including spaces and punctuation. Each prompt is assigned four fixed sampling seeds shared within its pair, yielding 2,400 images for a bilingual model and 1,200 for English-only evaluation.

\paragraph{Metric Computation.}
All metrics are computed from the outputs of \reward, which takes only the image as input, keeping recognition independent of the target text. For critical images, detected regions are matched to the requested targets with the many-to-many matching of Sec.~\ref{sec:method-reward}. CS averages the semantic score $s$ and CG the defect-free score $d$ of Eq.~\ref{eq:regional_reward} over all requested targets, with unmatched targets scored zero so that omitting text cannot reduce the denominator. For optional images, OG averages $d$ over the detected regions and then over images containing text, while text-free outputs are excluded rather than assigned a perfect score. All scores lie in $[0,1]$, with higher values indicating better quality.  Since \reward also serves as the training reward of \ourrl, we additionally conduct a user study as independent human validation, detailed in \ref{app:user_study}.

% \subsection{Additional Details on \reward}
% \label{supp:reward}
% \paragraph{Extreme Scene.}
% \paragraph{More Visualization.}
\subsection{Details of \reward-123K}
\label{supp:textdart-123K}

Existing text-centric corpora cover mostly clean, high-frequency text and annotate only readable content, providing no supervision for defect-aware transcription. \reward-123K is therefore built from complementary branches (Table~\ref{tab:textdart-sources}), including controlled prompt synthesis, caption-based regeneration, graphic design collections, long-tail and small-text supplementation, and text-free negatives. The controlled prompts follow the coverage dimensions in Table~\ref{tab:textdart-dimensions}. Images are rendered by multiple open-source and closed-source generators and are complemented by real-world imagery legally collected in-house, exposing rendering outcomes from clean text to severely distorted glyphs. Each image goes through OCR localization, region-wise TextPecker pre-annotation~\citep{textpecker}, human annotation, and two rounds of expert inspection, totaling approximately 12,500 person-hours. Annotations record the text actually rendered rather than the prompt intent~\citep{zhou2026flattery}. Every visible text piece is enclosed by a quadrilateral, transcriptions are literal with three structural-anomaly markers, and a readable but incorrect word is retained as rendered, since semantic comparison is performed separately by the regional reward. Text-free images receive empty region lists to supervise the empty-output case. We reserve 1,000 text-containing entries as the test split and use the remaining 122,200 entries for training, with disjoint sample IDs.
Table~\ref{tab:textdart-prompt} lists the instruction template of \reward, which mirrors the annotation protocol above. The template never exposes the requested target strings, consistent with the image-only assessment in Sec.~\ref{sec:method-reward}.
Table~\ref{tab:textdart-splits} summarizes the training split. Notably, 44.9\% of training entries mix marked and unmarked regions, providing direct supervision for regional quality differences that image-level rewards obscure. The test split contains 1,000 text-containing entries with 4,383 annotated regions, balanced by design across language (600 Chinese and 400 English) and anomaly presence (half anomalous per language).

\begin{table*}[t]
\centering
\caption{\textbf{Instruction Template of \reward} for defect-aware text localization and transcription.}
\label{tab:textdart-prompt}
\small
\setlength{\tabcolsep}{5pt}
\begin{tabular}{p{0.10\textwidth}p{0.84\textwidth}}
\toprule
Component & Content \\
\midrule
Role & You are an expert in perceiving structural anomalies of rendered text in AI-generated images. \\
Task & Given a generated image, locate and transcribe every piece of visible text, including distorted, broken, or entirely unreadable text, and report each piece as a quadrilateral region together with its transcription. \\
Marking & \texttt{<\#>} for a single structurally flawed and unrecognizable Chinese character or English letter.\newline
\texttt{<\#\#\#>} for a single structurally flawed and unrecognizable English word (never a phrase, line, or sentence).\newline
\texttt{<\#\#\#\#\#>} for an entirely unreadable secondary text block whose characters cannot be counted individually.\newline
Transcribe literally what is rendered: write out every readable character, never guess plausible words from context, never silently repair broken text, and keep Traditional or Simplified Chinese as rendered. Separate multiple lines within one quadrilateral by a single space. \\
Grounding & Every visible text piece is covered by at least one quadrilateral $[x_1, y_1, x_2, y_2, x_3, y_3, x_4, y_4]$, whose vertices start at the top-left corner in the semantic reading direction and proceed clockwise, including vertical, rotated, or curved text. Text that wraps over lines, curves, or is broken up by graphics is emitted as several quadrilaterals instead of one loose box. \\
Output & A single JSON object: \texttt{\{"regions": [\{"quad": [x1, y1, ..., x4, y4], "recognized\_text": "text with markers"\}]\}}. \\
\bottomrule
\end{tabular}
\end{table*}

\begin{table}[t]
\centering
\caption{\textbf{Composition of the \reward-123K Training Split.} The upper rows report counts and the lower rows report ratios. Image-level ratios use the 100,352 text-containing entries, and the region-level ratio uses all 495,778 annotated regions.}
\label{tab:textdart-splits}
\small
\setlength{\tabcolsep}{5pt}
\begin{tabular}{lc}
\toprule
Statistic & Train (122K) \\
\midrule
Sample entries & 122,200 \\
\quad Text-containing & 100,352 \\
\quad Text-free & 21,848 \\
Annotated regions & 495,778 \\
Entries with anomaly markers & 55,587 (55.4\%) \\
\midrule
Multi-region entries & 81.1\% \\
Entries with $\geq$6 regions & 25.0\% \\
Entries mixing marked and unmarked regions & 44.9\% \\
Regions with anomaly markers & 35.1\% \\
Artistic-text entries & 25.7\% \\
Script: CJK-only / Latin-only / mixed / digits \& symbol & 54.7 / 13.2 / 30.9 / 1.2\% \\
\bottomrule
\end{tabular}
\end{table}

\subsection{Discussion}
\paragraph{Suppression of hallucinated text under prompt enhancement.}
\reward penalizes glyph defects strictly, while the small text that generative models hallucinate spontaneously is almost always malformed. With prompt enhancement enabled, prompts become richer in detail and models are more prone to adding small text on their own. Such text is left unconstrained by recognition-based rewards and typically appears with malformed glyphs. Models trained with \ourrl instead tend to render the text requested by the prompt accurately, and even spontaneously added small text remains legible, so their advantage grows further under evaluation with prompt enhancement.

\paragraph{Limitations and future work.}
(1) Our reward currently evaluates only the content correctness and glyph quality of text, and does not assess text layout, that is, whether the position, size, and relative arrangement of text regions match the prompt. Since \reward natively outputs text quadrilaterals, incorporating layout consistency into the reward is a direct extension, for example by constructing layout rewards from the relative positions between predicted boxes and described targets, or by combining with box-conditioned generation~\citep{zhu2024multibooth} for controllable typesetting. (2) \reward{}-123K currently covers English and Chinese, and defect annotation and reward training for more languages remain to be explored. (3) The principle of blaming errant words rather than the whole canvas is not limited to text~\citep{Archer,ASPO,zhu2024instantswap,fang2024real}. Any localizable defect can be incorporated into the same framework of regional rewards and spatial credit assignment, and extending this framework to temporal credit assignment in video generation~\citep{ma2025controllable,ma2024follow,feng2026enhancing,feng2026narrlv,ling2025vmbench,fu2026videoawr} is a natural next step.

\end{document}